\ifdefined\pdfminorversion\pdfminorversion=7\fi
\documentclass[preprint,review,12pt,authoryear,nopreprintline]{elsarticle}
\usepackage[T1]{fontenc}
\usepackage[utf8]{inputenc}
\usepackage{lmodern}
\usepackage[table]{xcolor}
\usepackage{amsmath,amsfonts,amssymb,bm}
\usepackage{array,booktabs,multirow,makecell,tabularx}
\usepackage{textcomp,graphicx,url,float}
\newcommand{\manuscriptfigdir}{}
\IfFileExists{DCRF_alignment.pdf}{}{\renewcommand{\manuscriptfigdir}{MQCODet_ISPRS_revised/}}
\usepackage[font=footnotesize,labelfont=normalfont]{caption}
\usepackage[ruled,vlined,linesnumbered]{algorithm2e}
\usepackage{microtype}
\usepackage{placeins}
\definecolor{ISPRSlinkblue}{HTML}{0077B5}
\usepackage[colorlinks=true,linkcolor=ISPRSlinkblue,citecolor=ISPRSlinkblue,urlcolor=ISPRSlinkblue,bookmarksnumbered=true]{hyperref}
\AtBeginDocument{\hypersetup{linkcolor=ISPRSlinkblue,citecolor=ISPRSlinkblue,urlcolor=ISPRSlinkblue}}
\makeatletter
\renewcommand{\emailauthor}[2]{\stepcounter{ead}%
  \g@addto@macro\@elseads{\raggedright\let\corref\@gobble
    \eadsep\href{mailto:#1}{\nolinkurl{#1}} (#2)%
    \def\eadsep{\unskip,\space}}}
\makeatother
\hypersetup{pdfauthor={Hui Li, Jianbo Ma, Hui Luo, Guanglin Qi, Zhixing Wang, Haorui Zuo, Jianlin Zhang},pdftitle={MQCODet: Modality Quality-Guided Collaborative Optimization for UAV-Based RGB-T Tiny Pedestrian Detection}}
\biboptions{sort&compress}

\SetKwInput{KwIn}{Input}
\SetKwInput{KwOut}{Output}
\definecolor{captionyellow}{HTML}{E2B000}
\newcolumntype{Y}{>{\centering\arraybackslash}X}
\DeclareUnicodeCharacter{00B2}{\ensuremath{^{2}}}
\journal{ISPRS Journal of Photogrammetry and Remote Sensing}
\date{}
\begin{document}
\begin{frontmatter}
\title{MQCODet: Modality Quality-Guided Collaborative Optimization for UAV-Based RGB-T Tiny Pedestrian Detection}
\author[a,b]{Hui~Li\fnref{equal}}
\ead{lihui242@mails.ucas.ac.cn}
\author[a,b]{Jianbo~Ma\fnref{equal}}
\ead{majianbo22@mails.ucas.ac.cn}
\author[a]{Hui~Luo\corref{cor}}
\ead{luohui19@mails.ucas.ac.cn}
\author[a]{Guanglin~Qi}
\ead{qiguanglin@ioe.ac.cn}
\author[a]{Zhixing~Wang}
\ead{202111012048@std.uestc.edu.cn}
\author[a]{Haorui~Zuo}
\ead{zuohaorui@ioe.ac.cn}
\author[a]{Jianlin~Zhang\corref{cor}}
\ead{jlin@ioe.ac.cn}
\address[a]{State Key Laboratory of Optical Field Manipulation Science and Technology, Institute of Optics and Electronics, Chinese Academy of Sciences, Chengdu 610209, China}
\address[b]{School of Electronic, Electrical and Communication Engineering, University of Chinese Academy of Sciences, Beijing 100049, China}
\fntext[equal]{Hui Li and Jianbo Ma contributed equally to this work.}
\cortext[cor]{Corresponding authors: Hui Luo and Jianlin Zhang.}
\begin{abstract}
Unmanned aerial vehicle (UAV)-based RGB-thermal (RGB-T) remote sensing provides complementary visual and thermal information for pedestrian perception under complex aerial observation conditions. However, detecting tiny pedestrians in UAV-based RGB-T imagery remains challenging due to cross-modal spatial heterogeneity, manifested as modality quality imbalance and spatial position misalignment. Existing RGB-T detectors primarily address these challenges at the feature fusion stage, without establishing an effective coupling between feature fusion and object localization, potentially resulting in suboptimal detection performance. To address this limitation, we introduce a Modality Quality-Guided Collaborative Optimization Detector (MQCODet), which extends modality quality from a fusion prior to a localization supervision cue. Specifically, we propose a Dynamic Calibration and Region-Aware Fusion (DCRF) module that progressively refines cross-modal spatial alignment and performs region-aware fusion to generate reliable multi-scale features and a scene-level modality quality prior. We further design an Efficient Contextual Interaction (ECI) encoder to enhance multi-scale feature interaction while preserving fine-grained spatial details. Moreover, we propose a Modality-Aware Adaptive (MAA) loss that incorporates the modality prior and object scale to adaptively balance spatial consistency and geometric similarity for localization supervision.
Extensive experiments demonstrate that MQCODet surpasses state-of-the-art methods on two UAV-based RGB-T tiny object detection benchmarks while achieving real-time inference on the edge platform. 
\end{abstract}
\end{frontmatter}

\section{Introduction}
\label{sec:intro}

Unmanned aerial vehicle (UAV)-based RGB-T pedestrian detection has emerged as an important remote sensing task. UAV-based sensing systems provide flexible observation perspectives and broad spatial coverage, while RGB and thermal modalities offer complementary visual information. These advantages enable reliable pedestrian perception across diverse scenarios, including disaster rescue~\citep{zhang2025aerial}, traffic monitoring~\citep{sun2022dronevehicle,ma2025multi}, and public surveillance~\citep{du2018uav,zhu2018visdrone}. 
Despite its practical value and potential for all-day perception, UAV-based RGB-T pedestrian detection remains challenging under complex imaging conditions, where modality quality exhibits substantial variation across scenes and spatial regions.

\begin{figure}[htbp]
\centering
\includegraphics[width=\columnwidth]{\manuscriptfigdir 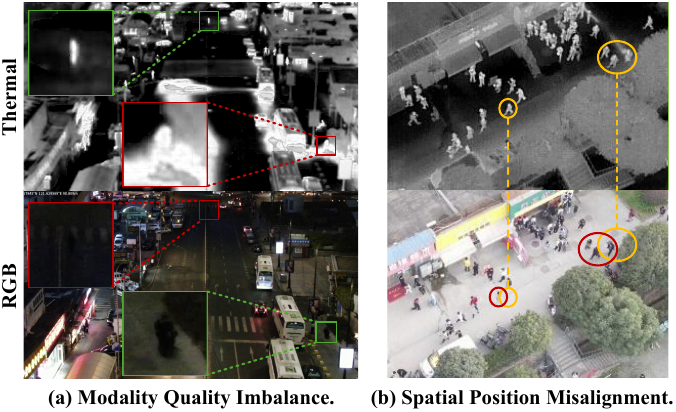}
\caption{Challenges in UAV-based RGB-T tiny pedestrian detection.
\textbf{(a)} Regional modality quality imbalance. Low illumination and thermal clutter cause spatially varying modality reliability within a scene. \textcolor{green}{Green} and \textcolor{red}{red} boxes denote reliable and unreliable regions, respectively.
\textbf{(b)} Spatial position misalignment. \textcolor{captionyellow}{Yellow} circles denote thermal pedestrian locations, and \textcolor{red}{red} circles denote RGB pedestrian locations. Spatial offsets impair cross-modal correspondence.}
  \label{fig:motivation}
\end{figure}

These variations stem from the distinct motion patterns of UAV platforms~\citep{ma2026tracking} and the different imaging mechanisms of RGB and thermal sensors~\citep{hwang2015multispectral}, resulting in pronounced cross-modal spatial heterogeneity that hinders reliable tiny pedestrian detection. 
This heterogeneity manifests as modality quality imbalance and spatial position misalignment. Here, modality quality refers to how reliably local RGB or thermal information supports pedestrian classification and localization. An imbalance arises when one modality provides more reliable evidence than the other within the same region. For example, the visible modality degrades in low-light regions, whereas the thermal modality becomes unreliable in areas with high background radiation (see Fig.~\ref{fig:motivation}(a)). 
Such spatially varying modality quality complicates cross-modal feature fusion, as unreliable responses may impair object representations. 
Furthermore, inherent viewpoint differences between RGB and thermal sensors lead to spatial misalignment of the same object across the paired images, which is particularly detrimental to tiny pedestrians with limited spatial details and high sensitivity to slight spatial displacement (see Fig.~\ref{fig:motivation}(b)).
Consequently, effectively mitigating cross-modal spatial heterogeneity is essential for UAV-based RGB-T tiny pedestrian detection under complex aerial conditions.

Existing RGB-T detection methods have explored different strategies to mitigate cross-modal spatial heterogeneity. 
Illumination-aware fusion methods~\citep{tinet,hu2025ei2det,Zhao_2025_ICCV} exploit environmental conditions to adaptively adjust the contributions of RGB and thermal modalities during fusion, improving robustness to global or local illumination changes. 
However, illumination cues alone cannot capture cross-modal spatial correspondence, potentially resulting in suboptimal fusion under spatial position misalignment. 
Alignment-aware fusion paradigms~\citep{yuan2024c2former,guo2024damsdet,peng2025coxnet} establish cross-modal spatial correspondence before or during feature interaction, mitigating the adverse effects of spatial position misalignment. 
Nevertheless, these paradigms focus on spatial correspondence without explicitly modeling spatially varying modality reliability, potentially limiting adaptive modality fusion. 
Recently, some works~\citep{zhang2023rgbtdroneperson,zhang2025c2dff} have explored quality-aware fusion strategies that incorporate regional modality quality to guide spatially adaptive fusion, enabling dynamic weighting of modality contributions. 
However, these methods primarily use modality quality to guide feature fusion, without propagating this reliability information to subsequent localization. 
This separation between quality-aware feature construction and quality-agnostic localization is particularly problematic for UAV-based RGB-T tiny pedestrian detection, where localization is sensitive to local modality reliability.

Building on this insight, we propose a Modality Quality-Guided Collaborative Optimization Detector (MQCODet) for UAV-based RGB-T tiny pedestrian detection. 
MQCODet is a unified detection framework that propagates modality quality from cross-modal feature fusion to localization supervision, while jointly enhancing cross-modal alignment and contextual feature interaction.
\textbf{First}, we propose a Dynamic Calibration and Region-Aware Fusion (DCRF) module to address modality quality imbalance and spatial position misalignment. 
DCRF progressively refines cross-modal spatial alignment and performs region-aware fusion to generate reliable multi-scale fused features and a scene-level modality prior. 
\textbf{Second}, we introduce an Efficient Contextual Interaction (ECI) encoder that enhances multi-scale feature interaction while preserving fine-grained spatial details through lightweight channel recalibration. 
\textbf{Third}, we introduce a Modality-Aware Adaptive (MAA) loss that leverages a scene-level modality prior to adaptively balance spatial consistency and geometric similarity for localization supervision.

\begin{figure}[htbp]
\centering\includegraphics[width=\columnwidth]{\manuscriptfigdir 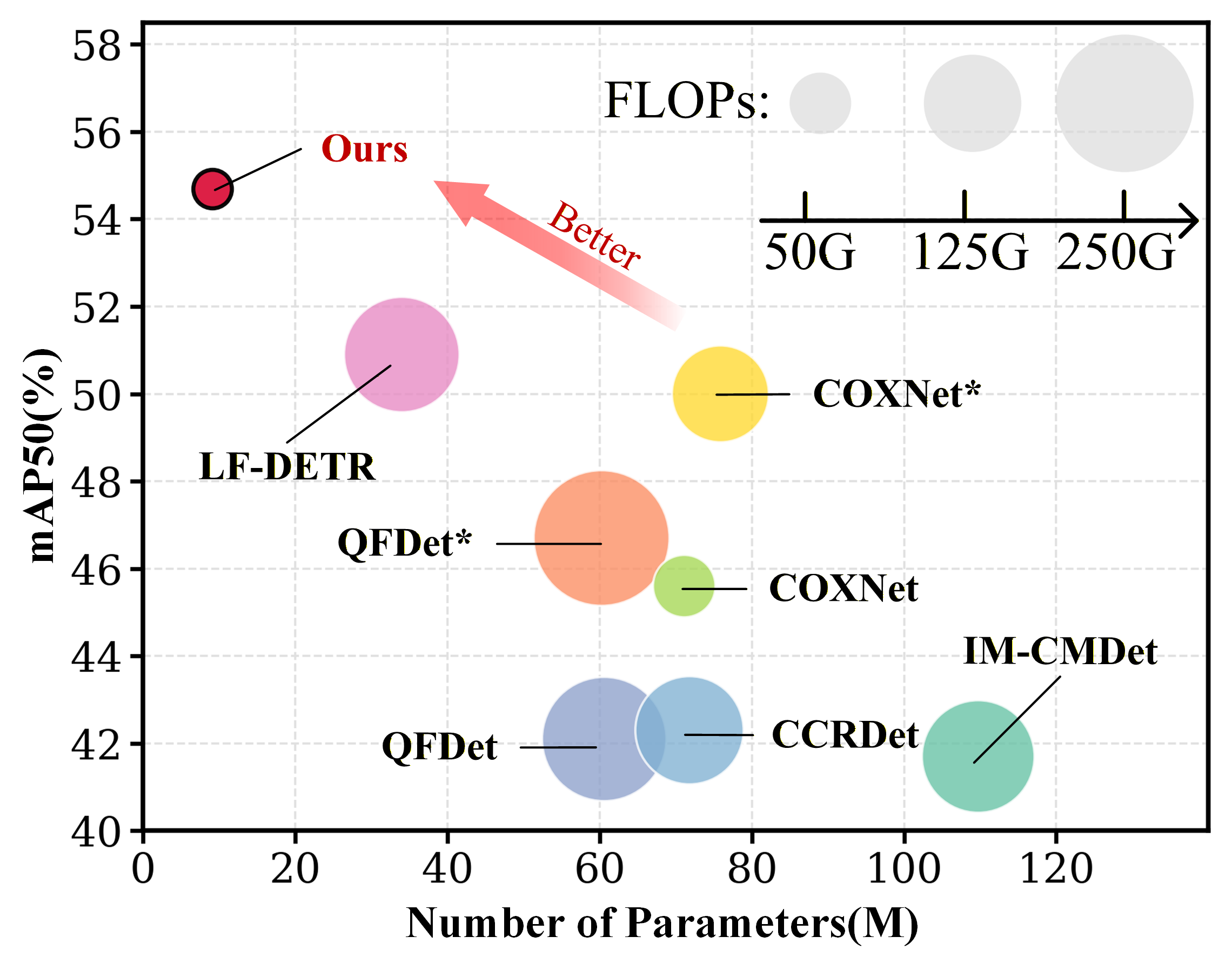}
\caption{Accuracy and complexity comparison on RGBTDronePerson. The vertical axis shows $mAP_{50}^{\mathrm{all}}$, the horizontal axis shows parameter count, and bubble area indicates FLOPs. MQCODet achieves the highest detection accuracy while offering a favorable trade-off between accuracy and computational complexity.}
\label{fig:bubble}
\end{figure}

As shown in Fig.~\ref{fig:bubble}, MQCODet achieves a favorable accuracy-efficiency trade-off on RGBTDronePerson, demonstrating its effectiveness while maintaining computational efficiency. 
The main contributions can be summarized as follows:
\begin{enumerate}
\item We propose a modality quality-guided collaborative optimization paradigm that extends modality quality from a fusion prior to a localization supervision cue, unifying quality-aware feature fusion and reliability-aware localization for UAV-based RGB-T tiny pedestrian detection.
\item We develop a Dynamic Calibration and Region-Aware Fusion module with an Efficient Contextual Interaction encoder to refine cross-modal spatial alignment, enhance regional feature fusion, and preserve fine-grained details in multi-scale representations.
\item We introduce a modality-aware adaptive loss that uses a scene-level modality quality prior and object scale to balance spatial consistency and geometric similarity for accurate pedestrian localization.
\end{enumerate}

\section{Related Work}
\label{sec:related}

\subsection{UAV Object Detection}
UAV imagery contains dense pedestrians with small spatial extent, large scale variation, and cluttered backgrounds~\citep{ma2024stcmot}. The UAV benchmark~\citep{du2018uav} and VisDrone~\citep{zhu2018visdrone} established standard testbeds for detection under diverse aerial conditions. Early detectors improved aerial representations through feature pyramids~\citep{lin2017fpn}, class-balanced supervision~\citep{lin2017focal}, and adaptive spatial sampling~\citep{dai2017deformable, 10653748}. Later studies strengthened tiny object features with sparse high-resolution queries~\citep{yang2022querydet} or auxiliary super-resolution learning~\citep{zhang2023superyolo}. The listed methods improve scale handling and spatial detail, but visible imagery alone cannot support reliable perception under poor illumination.
Recent UAV detectors exploit RGB and thermal observations to improve all-day perception. COXNet~\citep{peng2025coxnet} couples adaptive alignment with cross-level scale integration. IM-CMDet~\citep{luo2025imcmdet} combines intramodal enhancement and cross-modal fusion. Dynamic gated fusion~\citep{gu2025uav,li2025elnet} adjusts multimodal feature aggregation, while LF-DETR~\citep{qi2026lfdetr} strengthens frequency cues within a DETR detector. C2DFF-Net further combines spatial and frequency information within a multimodal remote-sensing detector~\citep{zhang2025c2dff}. HyperDet~\citep{li2026hyperdet} further models higher-order relations across modalities. 
Existing UAV detectors treat alignment, scale interaction, and
localization as independent components.
Such decoupled optimization prevents modality reliability from guiding the detection process.

\subsection{RGB-T Object Detection}
RGB-T object detection exploits color and texture from RGB images and thermal responses under limited illumination. The KAIST benchmark~\citep{hwang2015multispectral} supported early multispectral pedestrian detection, while cross-modal representation learning~\citep{xu2017learning} established joint feature learning across both modalities. \citet{cao2019boxlevel} used segmentation masks derived from bounding boxes to supervise a two-stream visible--thermal pedestrian detector. More recently, a visible--thermal tiny object benchmark has extended multispectral evaluation to the challenging small-object regime~\citep{ying2025visible}. Illumination-aware detection~\citep{li2019illumination} adjusted modality contributions with a global scene estimate. Shadows, thermal clutter, and uneven radiation create local quality variations beyond global scene estimates. Region-dependent degradation therefore requires spatially resolved modality assessment.
Cross-modal misalignment creates another source of unreliable evidence. Weakly aligned learning~\citep{zhang2019weakly} reduced dependence on accurate registration. C$^2$Former~\citep{yuan2024c2former} learned calibrated correspondence with cross-attention, while adaptive offset learning~\citep{chen2024weakly} and reflectance-guided alignment~\citep{zhao2025rgpfa} corrected spatial offsets in UAV imagery. Alignment improves feature correspondence but does not determine the reliability of aligned responses. A reliable detector must jointly estimate spatial correspondence and regional modality quality.
Feature fusion methods select complementary responses after feature extraction. GAFF~\citep{zhang2021gaff} uses guided attention, and ICAFusion~\citep{shen2024icafusion} performs iterative cross attention. GM-DETR~\citep{xiao2024gmdetr} introduces an efficient fusion encoder for end-to-end detection. MGFM~\citep{wang2024mgmf} and WaveMamba~\citep{zhu2025wavemamba} model cross-modal dependencies with state-space architectures. Removal-then-selection~\citep{zhao2025removal} suppresses degraded information before feature selection. For UAV-based tiny pedestrian detection, QFDet~\citep{zhang2023rgbtdroneperson} and CCRDet~\citep{gu2026ccrdet} use regional scene cues to improve complementary fusion. Existing fusion methods confine modality quality to feature aggregation. Independent localization losses ignore the reliability of multimodal evidence.

\subsection{Tiny Object Localization}
Tiny object localization requires stable supervision because a small coordinate shift can produce a large relative localization error. Under single-point supervision, Point Teacher~\citep{zhu2025pointteacher} refines pseudo boxes to address scale ambiguity and shifts in annotated point locations. For box-supervised detectors, IoU-based losses optimize geometric overlap. GIoU~\citep{rezatofighi2019generalized} extends overlap supervision to disjoint boxes, while DIoU~\citep{zheng2020distance} penalizes the squared distance between box centers normalized by the squared diagonal of the smallest enclosing box. Overlap-based objectives remain sensitive to pixel perturbations when a pedestrian occupies a few pixels. NWD~\citep{xu2022detecting} represents boxes as Gaussian distributions and provides smoother supervision for aerial tiny objects. Deformation sensitive localization~\citep{tang2025learning} links RGB-T fusion with geometric variation. Distribution-based objectives improve stability for tiny boxes, but weaker boundary constraints can reduce localization precision for larger boxes.
DETR variants improve localization through query design and distribution refinement. Deformable DETR~\citep{zhu2020deformable} samples sparse multiscale features. DAB-DETR~\citep{liu2022dabdetr} formulates queries as dynamic anchor boxes, and DINO~\citep{zhang2022dino} improves query initialization and denoising. RT-DETR~\citep{zhao2024rtdetr} develops an efficient hybrid encoder for real-time detection. D-FINE~\citep{peng2025dfine} refines box distributions, while DEIM~\citep{huang2025deim} improves matching quality and convergence. Localization objectives apply fixed regression rules across object scales and imaging conditions. Fixed rules cannot account for scale-dependent geometry or region-dependent modality ambiguity.

\begingroup
\section{Methodology}
\label{sec:method}

MQCODet uses modality quality to guide both feature fusion and localization supervision. DCRF aligns and fuses RGB and thermal features, ECI exchanges information across feature levels, and MAA adjusts the localization objective during training. Throughout this section, $l\in\{2,3,4,5\}$ denotes the feature level, and $m\in\{\mathrm{rgb},\mathrm t\}$ denotes the modality. Superscripts identify the source or processing stage of a quantity; image and batch indices are omitted when the context is clear.

\subsection{Overall Architecture}
\label{subsec:method_overall}

Fig.~\ref{fig:framework}(a) presents the overall architecture. Given paired RGB and thermal images $I^{\mathrm{rgb}},I^{\mathrm t}\in\mathbb R^{3\times H\times W}$, MQCODet uses two independent HGNetv2-B0 backbones initialized from the same pretrained weights. The two branches maintain separate trainable parameters during joint optimization. Here, $H$ and $W$ denote the input height and width, and the thermal input is represented using three channels. Each branch extracts four feature levels and projects them to a common channel width:
\begin{equation}
F_l^m=\operatorname{Backbone}_l^m(I^m).
\label{eq:features}
\end{equation}
The mapping $\operatorname{Backbone}_l^m$ includes feature extraction up to level $l$ and its subsequent $1\times1$ convolution with batch normalization. These output projections are independent across modalities and levels. The resulting feature $F_l^m\in\mathbb R^{C_f\times H_l\times W_l}$ has $C_f=192$ channels, where $H_l=H/2^l$ and $W_l=W/2^l$.

DCRF aligns RGB features to the thermal reference and produces fused features $F_l^{\mathrm{fuse}}$ according to regional modality reliability. ECI then exchanges contextual information through top-down and bottom-up paths, yielding the enriched features $E_l$ for the query decoder. The decoder predicts object categories and bounding boxes. During training, MAA combines an image-level quality prior from DCRF with object scale and boundary-distribution ranking to balance spatial consistency and geometric overlap.

\begin{figure}[htbp]
\centering
\includegraphics[width=0.98\textwidth]{\manuscriptfigdir 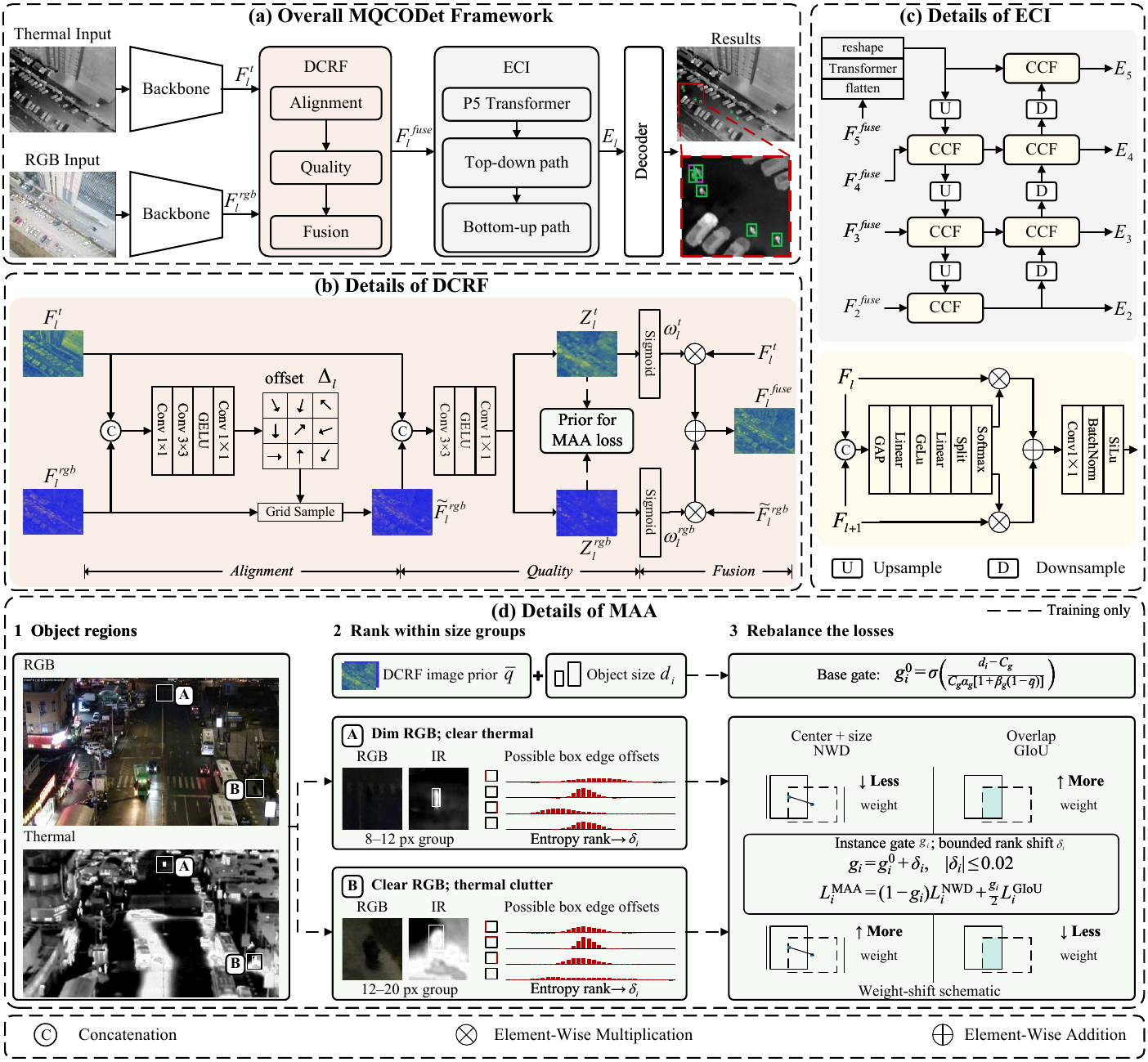}
\caption{Overview of MQCODet. (a) Independent RGB and thermal backbones produce features $F_l^{\mathrm{rgb}}$ and $F_l^{\mathrm t}$. DCRF generates fused features $F_l^{\mathrm{fuse}}$, and ECI produces enriched features $E_l$ for detection. (b) DCRF aligns RGB features to the thermal reference and estimates regional modality reliability for fusion. (c) ECI exchanges contextual information through top-down and bottom-up paths with context compensation fusion (CCF). (d) During training, the image-level quality prior $\bar q$ and object scale $d_i$ determine a base gate $g_i^0$, which is adjusted by a bounded ranking residual $\delta_i$ to obtain the localization gate $g_i$.}
\label{fig:framework}
\end{figure}

\subsection{Dynamic Calibration and Region-Aware Fusion}
\label{subsec:dcrf}

DCRF first establishes local spatial correspondence and then determines the contribution of each modality to fusion. Here, modality quality refers to the reliability of local modality information for pedestrian classification and localization.

As shown in Fig.~\ref{fig:framework}(b), the offset branch first projects both modality features to a lower channel width:
\begin{equation}
F_l^{m,\mathrm{proj}}=\operatorname{Proj}_l(F_l^m).
\label{eq:offset_projection}
\end{equation}
Here, the superscript $\mathrm{proj}$ denotes the projected feature. At each level, $\operatorname{Proj}_l$ is a $1\times1$ convolution shared by the two inputs to the offset branch, reducing the channel width from 192 to 32.

The projected features and their absolute difference are then concatenated to predict the spatial offsets:
\begin{equation}
\Delta_l=2\tanh\!\left(\operatorname{Offset}_l\!\left[F_l^{\mathrm t,\mathrm{proj}},F_l^{\mathrm{rgb},\mathrm{proj}},\left|F_l^{\mathrm t,\mathrm{proj}}-F_l^{\mathrm{rgb},\mathrm{proj}}\right|\right]\right).
\label{eq:offset}
\end{equation}
Here, brackets denote channel-wise concatenation, and $|\cdot|$ denotes element-wise absolute value. The offset predictor $\operatorname{Offset}_l$ uses depthwise and pointwise convolutions with GELU activations. Its output $\Delta_l\in\mathbb R^{2\times H_l\times W_l}$ contains horizontal and vertical offsets. The factor 2 and the hyperbolic tangent bound each component within $[-2,2]$ feature-grid cells.

Using the thermal grid as the reference, DCRF samples the RGB feature at the predicted locations:
\begin{equation}
\widetilde F_l^{\mathrm{rgb}}(\mathbf x)=\operatorname{GridSample}\!\left(F_l^{\mathrm{rgb}},\mathbf x+\Delta_l(\mathbf x)\right).
\label{eq:warp}
\end{equation}
Here, $\mathbf x$ is a reference-grid location, and $\operatorname{GridSample}$ denotes bilinear interpolation at the specified feature-grid coordinates. The output $\widetilde F_l^{\mathrm{rgb}}$ is the aligned RGB feature. The final offset layer is initialized to zero, so alignment starts from zero displacement.

The aligned features are then used to predict modality reliability:
\begin{equation}
Z_l=\operatorname{Quality}_l\!\left([\widetilde F_l^{\mathrm{rgb}},F_l^{\mathrm t}]\right).
\label{eq:reliability}
\end{equation}
The estimator $\operatorname{Quality}_l$ contains pointwise, depthwise, and pointwise convolutions with an intermediate width of 32. Its two-channel output $Z_l\in\mathbb R^{2\times H_l\times W_l}$ contains one raw response, or logit, $Z_l^m$ for each modality. The sigmoid function $\sigma$ converts each response into a local reliability score. These scores are normalized across modalities to obtain the spatial fusion weights:
\begin{equation}
\omega_l^m=\frac{\sigma(Z_l^m)}{\max\!\left(\sigma(Z_l^{\mathrm{rgb}})+\sigma(Z_l^{\mathrm t}),\varepsilon\right)}.
\label{eq:modality_weights}
\end{equation}
\begin{equation}
F_l^{\mathrm{fuse}}=\omega_l^{\mathrm{rgb}}\odot\widetilde F_l^{\mathrm{rgb}}+\omega_l^{\mathrm t}\odot F_l^{\mathrm t}.
\label{eq:fusion}
\end{equation}
Here, $\omega_l^m$ is the relative contribution of modality $m$ at each location, $\odot$ denotes element-wise multiplication with spatial weights broadcast across channels, and $\varepsilon=10^{-6}$ ensures numerical stability. The superscript $\mathrm{fuse}$ denotes the fused feature. Zero initialization of the final reliability layer gives equal initial modality weights. The task-derived supervision in Eq.~\eqref{eq:quality_teacher} connects the predicted reliability to local classification and localization performance.

\subsection{Efficient Contextual Interaction}
\label{subsec:eci}

As shown in Fig.~\ref{fig:framework}(c), ECI exchanges information between coarse semantic features and fine spatial features. Global context is first introduced at the deepest level:
\begin{equation}
F_5^{\mathrm{td}}=\operatorname{Transformer}(F_5^{\mathrm{fuse}}).
\label{eq:global_context}
\end{equation}
The superscript $\mathrm{td}$ denotes a feature in the top-down path. The Transformer operation includes flattening, one encoder layer with two-dimensional positional embeddings, and reshaping back to the original spatial dimensions. It uses eight attention heads and a feed-forward width of 512.

At each cross-level connection, context compensation fusion (CCF) determines how much each input contributes in each channel. Let $F^{(1)}$ and $F^{(2)}$ denote the two inputs after their spatial and channel dimensions have been matched. Node indices are omitted in the following CCF equations. The channel weights are
\begin{equation}
\left[\omega^{\mathrm{ctx},1};\,\omega^{\mathrm{ctx},2}\right]=\operatorname{Softmax}_{\mathrm{branch}}\!\left(\operatorname{MLP}\!\left[\operatorname{GAP}(F^{(1)}),\operatorname{GAP}(F^{(2)})\right]\right).
\label{eq:context_weights}
\end{equation}
Global average pooling, denoted by $\operatorname{GAP}$, averages each channel over its spatial positions. The operation $\operatorname{MLP}$ maps the concatenated descriptor through dimensions $384\rightarrow24\rightarrow384$ with a GELU activation, then reshapes the result into two rows of length $C_f$. The semicolon on the left denotes vertical stacking of the two row vectors. The softmax normalizes the two rows for each channel. The resulting row vectors $\omega^{\mathrm{ctx},1}$ and $\omega^{\mathrm{ctx},2}$ are channel weights; the tag $\mathrm{ctx}$ distinguishes them from the spatial modality weights in DCRF. CCF then produces
\begin{equation}
\operatorname{CCF}(F^{(1)},F^{(2)})=2\left(\omega^{\mathrm{ctx},1}\odot F^{(1)}+\omega^{\mathrm{ctx},2}\odot F^{(2)}\right).
\label{eq:context_fusion}
\end{equation}
The channel weights are broadcast over spatial positions. The final linear layer is initialized to zero, giving both weights the initial value $1/2$. The factor 2 therefore makes the initial output equal to $F^{(1)}+F^{(2)}$.

Starting from $F_5^{\mathrm{td}}$, the top-down path successively processes levels $l=4,3,2$:
\begin{equation}
F_{l+1}^{\mathrm{lat}}=\operatorname{Proj}_{l+1}^{\mathrm{lat}}(F_{l+1}^{\mathrm{td}}).
\label{eq:lateral_projection}
\end{equation}
\begin{equation}
F_l^{\mathrm{td}}=\operatorname{Agg}_l^{\mathrm{td}}\!\left(\operatorname{CCF}_l^{\mathrm{td}}\!\left(\operatorname{Up}(F_{l+1}^{\mathrm{lat}}),F_l^{\mathrm{fuse}}\right)\right).
\label{eq:topdown}
\end{equation}
The superscript $\mathrm{lat}$ identifies a lateral feature produced by a convolution and normalization. These features are retained for the bottom-up path. The operator $\operatorname{Up}$ denotes nearest-neighbor upsampling, and $\operatorname{Agg}$ denotes the convolutional aggregation block following CCF.

The bottom-up path starts from $E_2=F_2^{\mathrm{td}}$ and processes levels $l=3,4,5$:
\begin{equation}
E_l=\operatorname{Agg}_l^{\mathrm{bu}}\!\left(\operatorname{CCF}_l^{\mathrm{bu}}\!\left(\operatorname{Down}_{l-1}(E_{l-1}),F_l^{\mathrm{lat}}\right)\right).
\label{eq:bottomup}
\end{equation}
Here, $\operatorname{Down}_{l-1}$ downsamples the preceding output to the current resolution, and $\mathrm{bu}$ identifies the bottom-up path. The six CCF nodes have separate parameters. The enriched outputs $\{E_2,E_3,E_4,E_5\}$ are passed to the query decoder.

\subsection{Modality-Aware Adaptive Loss}
\label{subsec:maa}

For tiny objects, even small localization errors can lead to substantial changes in overlap. Moreover, objects of similar size may exhibit different boundary-prediction distributions. To account for these variations, the MAA loss first constructs a base gate based on image-level quality and object scale, then refines it using boundary-distribution ranks within each size group, as illustrated in Fig.~\ref{fig:framework}(d).

\subsubsection{Regional quality and image-level prior}
For an image containing $N$ annotated objects, let $b_i=(c_i^x,c_i^y,w_i,h_i)$ denote the normalized ground-truth bounding box of object $i\in\{1,\ldots,N\}$, where $(c_i^x,c_i^y)$ represents its center coordinates, and $w_i$ and $h_i$ denote its width and height, respectively. We extract the DCRF logits within each target region and aggregate the resulting quality estimates as follows:
\begin{equation}
q_{i,l}^m=\sigma\!\left(\operatorname{ROI}_3(Z_l^m,b_i)\right).
\label{eq:regional_quality}
\end{equation}
\begin{equation}
\bar q=\frac{1}{4N}\sum_{i=1}^{N}\sum_{l=2}^{5}\max_m q_{i,l}^m.
\label{eq:anchor_direct}
\end{equation}
The operator $\operatorname{ROI}_3$ applies $3\times3$ ROIAlign~\citep{he2017mask} to each target region and averages the nine sampled logits before applying the sigmoid function, yielding the regional quality estimate $q_{i,l}^{m}$. For each target at each feature level, we retain the stronger quality response by taking the maximum across modalities. These responses are then averaged over feature levels and targets to obtain the image-level prior $\bar q$. The resulting prior summarizes the reliability available within target regions and serves as a shared quality cue for all object gates in the image.
For images without annotated objects, no object gates are computed.

We measure object scale in a $640\times640$ reference coordinate system and define the base gate $g_i^0$ as: 
\begin{equation}
g_i^0=\sigma\!\left(\frac{d_i-C_g}{C_g\alpha_g[1+\beta_g(1-\operatorname{sg}(\bar q))]}\right).
\label{eq:base_gate}
\end{equation}
Here, $d_i=\sqrt{(640w_i)(640h_i)}$ is the equivalent target side length, and $\operatorname{sg}$  denotes the stop-gradient operator. 
The transition scale is $C_g=32$, while $\alpha_g=1$ and $\beta_g=1$ control the base temperature and quality modulation. 
The resulting gate favors spatial-distance supervision for smaller objects and geometric-overlap supervision for larger objects.
As image-level quality decreases, the effective temperature increases, moving $g_i^0$ toward $1/2$ and thereby moderating this scale-dependent preference.

\subsubsection{Instance-level adjustment}
For each box edge, the decoder predicts a distribution over 33 discrete locations.
Let $Z_{i,a,k}^{\mathrm{loc}}$ denote the localization logit produced by the final-layer query assigned to object $i$ via Hungarian matching, where $a\in\{1,2,3,4\}$ indexes the left, top, right, and bottom edges, respectively, and $k\in\{0,\ldots,32\}$ indexes the discrete locations.
We obtain the boundary probabilities by applying softmax over the discrete locations of each edge to the detached logits:
\begin{equation}
p_{i,a,k}=\operatorname{Softmax}_k\!\left(\operatorname{sg}(Z_{i,a,k}^{\mathrm{loc}})\right).
\label{eq:boundary_probabilities}
\end{equation}
Using these probabilities, we define an entropy-based ranking score as the mean negative entropy across the four boundary distributions:
\begin{equation}
r_i^{\mathrm{ent}}=\frac14\sum_{a=1}^{4}\sum_{k=0}^{32}p_{i,a,k}\log p_{i,a,k}.
\label{eq:negative_entropy}
\end{equation}
Higher values of $r_i^{\mathrm{ent}}$ indicate more concentrated boundary distributions.
We use this score as a relative ranking signal rather than a calibrated measure of localization accuracy.

To compare boundary predictions at similar scales, we partition the targets within each image into five groups according to $d_i$: $(0,8]$, $(8,12]$, $(12,20]$, $(20,32]$, and $(32,+\infty)$.
Let $\mathcal G$ denote a size group containing $N_{\mathcal G}$ targets.
We apply rank-based refinement only when $N_{\mathcal G}\geq3$, all targets in the group have matched final-layer queries, and their ranking scores are pairwise distinct.
For each eligible group, we sort $r_i^{\mathrm{ent}}$ in ascending order and assign integer ranks $r_i\in\{0,\ldots,N_{\mathcal G}-1\}$, with higher ranks corresponding to more concentrated boundary distributions. 
To bound the refinement while keeping the adjusted gates within $[0,1]$, we define the group-specific adjustment limit as:
\begin{equation}
\delta_{\mathcal G}^{\max}
=
\min\!\left(
\eta,
\min_{i'\in\mathcal G}
\min(g_{i'}^0,1-g_{i'}^0)
\right),
\label{eq:rank_bound}
\end{equation}
where $i'$ indexes the targets in $\mathcal G$, and $\eta\geq0$ controls the maximum adjustment magnitude.
Unless otherwise specified, we set $\eta=0.02$.
We then map each rank to a symmetric residual and add it to the corresponding base gate:
\begin{equation}
\delta_i
=
\delta_{\mathcal G}^{\max}
\left(
\frac{2r_i}{N_{\mathcal G}-1}-1
\right),
\label{eq:rank_residual}
\end{equation}
\begin{equation}
g_i=g_i^0+\delta_i,
\label{eq:rank_gate}
\end{equation}
where $\delta_i$ is the ranking residual and $g_i$ is the final gate.
For groups that do not satisfy the eligibility conditions, we set $\delta_i=0$, leaving the base gates unchanged. 
This refinement guarantees $|\delta_i|\leq\eta$ and $g_i\in[0,1]$.
The symmetric rank mapping also ensures $\sum_{i\in\mathcal G}\delta_i=0$, preserving the group's mean gate value.
Thus, higher-ranked targets receive larger residuals relative to their respective base gates, shifting supervision toward geometric overlap without changing the group's average weighting.

\subsubsection{Mixed localization objective}
For a normalized box prediction $\widehat b$ matched to target $i$, we combine spatial-distance and overlap-based supervision using the normalized Wasserstein distance (NWD) loss~\citep{xu2022detecting} and the generalized intersection over union (GIoU) loss~\citep{rezatofighi2019generalized}.
To compute NWD in the $640\times640$ reference coordinate system, we define
$b_i^{\mathrm{NWD}}=640(c_i^x,c_i^y,w_i/2,h_i/2)$
and construct $\widehat b^{\mathrm{NWD}}$ from the predicted coordinates in the same way.
These representations encode the box centers and half-size coordinates in the reference system.
The two losses are then given by: 
\begin{equation}
\mathcal L_i^{\mathrm{NWD}}(\widehat b)=1-\exp\!\left(-\frac{\|\widehat b^{\mathrm{NWD}}-b_i^{\mathrm{NWD}}\|_2}{32}\right),
\label{eq:similarities}
\end{equation}
\begin{equation}
\mathcal L_i^{\mathrm{GIoU}}(\widehat b)=1-\operatorname{GIoU}(\widehat b,b_i),
\label{eq:giou_loss}
\end{equation}
where $\|\cdot\|_2$ denotes the Euclidean norm, $\exp$ denotes the exponential function, and $32$ is the NWD normalization constant in reference pixels.
Unlike NWD, GIoU can be computed directly from the normalized boxes because it is invariant to the common coordinate scaling.

Using the final target gate $g_i$, we define the object-level MAA loss as: 
\begin{equation}
\mathcal L_i^{\mathrm{MAA}}(\widehat b)=(1-g_i)\mathcal L_i^{\mathrm{NWD}}(\widehat b)+\frac{g_i}{2}\mathcal L_i^{\mathrm{GIoU}}(\widehat b).
\label{eq:maa_object_loss}
\end{equation}
The factor $1/2$ rescales the GIoU loss so that both loss terms lie within $[0,1]$. For prediction branch $s$, we aggregate the object-level losses over its matched prediction--target pairs:
\begin{equation}
\mathcal L_{\mathrm{MAA}}^{(s)}=\frac{1}{N_s}\sum_{(j,i)\in\Omega_s}\mathcal L_i^{\mathrm{MAA}}(\widehat b_j^{(s)}),
\label{eq:maa_loss}
\end{equation}
where $s$ indexes prediction branches, $j$ indexes predictions, $\Omega_s$ denotes the set of matched prediction--target pairs for branch $s$, and $N_s>0$ is its original normalization factor.

Each target gate is computed once and reused across the main, auxiliary, encoder, and denoising branches according to their assigned target indices.
Each branch retains its own matching procedure and normalization factor, including repeated assignments to the same target where applicable.
Gate construction is detached from gradient computation, while localization gradients propagate through the predicted boxes.

\subsubsection{Reliability supervision and joint training}
We supervise the modality reliability estimates using local task probes.
At each feature level, a modality-specific probe predicts class logits and bounding boxes from the corresponding aligned features. 
For the grid location assigned to target $i$, let $Z_{i,l}^{m,\mathrm{cls}}$ denote the logit for the ground-truth category and $\widehat b_{i,l}^{m,\mathrm{probe}}$ the normalized box prediction.
The superscripts $\mathrm{cls}$ and $\mathrm{probe}$ distinguish the classification and probe outputs, respectively. 
We define the classification and localization errors as: 
\begin{equation}
e_{i,l}^{m,\mathrm{cls}}=-\log\sigma(Z_{i,l}^{m,\mathrm{cls}}).
\label{eq:probe_classification_error}
\end{equation}
\begin{equation}
e_{i,l}^{m,\mathrm{box}}=\operatorname{Mean}\!\left[\log\!\left(1+\frac{|\widehat b_{i,l}^{m,\mathrm{probe}}-b_i|}{\max((w_i,h_i,w_i,h_i),1/640)}\right)\right].
\label{eq:probe_errors}
\end{equation}
Here, $\operatorname{Mean}$ averages the four box-coordinate terms. The maximum and the division are applied element-wise.
We combine the classification and localization errors to obtain the quality supervision target
\begin{equation}
q_{i,l}^{m,\mathrm{sup}}=\exp\!\left[-\operatorname{sg}\!\left(e_{i,l}^{m,\mathrm{cls}}+e_{i,l}^{m,\mathrm{box}}\right)\right].
\label{eq:quality_teacher}
\end{equation}
Here, the superscript $\mathrm{sup}$ identifies the supervision target. 
Smaller classification and localization errors produce a higher target score, tying modality reliability to local detection performance.

Using these targets, we train the reliability estimators and task probes with
\begin{equation}
\mathcal L_{\mathrm{rel}}=\left\langle\operatorname{BCELogits}\!\left(\operatorname{ROI}_3(Z_l^m,b_i),q_{i,l}^{m,\mathrm{sup}}\right)\right\rangle_{i,l,m}.
\label{eq:reliability_loss}
\end{equation}
\begin{equation}
\mathcal L_{\mathrm{probe}}=\frac18\sum_{l=2}^{5}\sum_m\left(\mathcal L_{\mathrm{focal}}^{m,l}+\left\langle e_{i,l}^{m,\mathrm{box}}\right\rangle_i\right).
\label{eq:auxiliary_losses}
\end{equation}
The operator $\operatorname{BCELogits}$ denotes binary cross-entropy computed from logits, and angle brackets denote averages over the indicated indices and valid targets across the mini-batch.
An average over an empty target set contributes zero.
The factor $1/8$ averages the probe losses over the two modalities and four feature levels. 
For dense classification, $\mathcal L_{\mathrm{focal}}^{m,l}$ is the standard focal loss~\citep{lin2017focal}, with positive-class weight $0.25$ and focusing exponent $2$.
At each target-center location, the entry corresponding to the ground-truth category is positive, while all remaining grid--class entries are negative.

The detection objective combines the losses supported by each prediction branch: 
\begin{equation}
\mathcal L_{\mathrm{det}}=\sum_s\left(\mathcal L_{\mathrm{MAL}}^{(s)}+5\mathcal L_1^{(s)}+0.15\mathcal L_{\mathrm{FGL}}^{(s)}+1.5\mathcal L_{\mathrm{DDF}}^{(s)}+2\mathcal L_{\mathrm{MAA}}^{(s)}\right).
\label{eq:detection_objective}
\end{equation}
Here, the Matchability-Aware Loss $\mathcal L_{\mathrm{MAL}}$ supervises classification~\citep{huang2025deim}, $\mathcal L_1$ denotes the bounding-box $L_1$ loss, and $\mathcal L_{\mathrm{FGL}}$ and $\mathcal L_{\mathrm{DDF}}$ denote the Fine-Grained Localization and Decoupled Distillation Focal losses, respectively~\citep{peng2025dfine}.
Each branch contributes only the loss terms supported by its outputs.

Finally, we incorporate the probe and reliability losses into the complete training objective:
\begin{equation}
\mathcal L_{\mathrm{total}}=\mathcal L_{\mathrm{det}}+0.1\left(\mathcal L_{\mathrm{probe}}+\mathcal L_{\mathrm{rel}}\right).
\label{eq:training_objective}
\end{equation}
During inference, the backbones, DCRF, ECI, and decoder produce detections.
The task probes, annotated-region readout, and MAA gate construction are used only during training and are omitted from the inference pipeline.
\endgroup

\section{Experiments}
\label{sec:experiments}

\begingroup

\subsection{Experimental Setup}
\label{subsec:exp_setup}

\subsubsection{Datasets}
We evaluate MQCODet on three UAV-based RGB-T object detection benchmarks: RGBTDronePerson~\citep{zhang2023rgbtdroneperson}, VTUAV-det~\citep{zhang2023rgbtdroneperson,zhang2022vtuav}, and DroneVehicle~\citep{sun2022dronevehicle}. All methods are trained separately on each dataset.

\noindent\textbf{RGBTDronePerson.}
This dataset contains 6,125 RGB-T image pairs collected under diverse illumination and thermal conditions. The annotations include person, rider, crowd, and uncertain labels, with uncertain instances excluded from training and evaluation under the benchmark protocol. It features tiny targets and spatially varying modality quality. Following the established split, we use 4,900 image pairs for training and 1,225 for validation, and report detection performance on the validation split.

\noindent\textbf{VTUAV-det.}
VTUAV-det is a pedestrian detection benchmark derived from VTUAV, a visible-thermal UAV tracking dataset. It contains 16,770 RGB-T image pairs and considers only the person category. The dataset covers diverse scenes and weather conditions, with substantial variation in object scale. Following the benchmark split, 11,392 image pairs are used for training and 5,378 for testing.

\noindent\textbf{DroneVehicle.}
DroneVehicle contains 28,439 RGB-T image pairs annotated with five vehicle categories: car, truck, bus, van, and freight car. The dataset covers vehicles of different shapes and scales across diverse road scenes and thermal conditions. The official split comprises 17,990 training pairs, 1,469 validation pairs, and 8,980 test pairs. In this work, all methods are trained on the official training split and evaluated on the validation split.

\subsubsection{Compared Methods}

The superscript $\dagger$ denotes an RGB-T adaptation of a detector
originally designed for single-modality input.
D-FINE$^\dagger$~\citep{peng2025dfine} and
DEIM$^\dagger$~\citep{huang2025deim} both adopt a dual-branch
architecture with independent RGB and thermal backbones, using
HGNet and HGNetv2-B0, respectively. Features from the two branches
are fused at corresponding levels before being passed to the encoder. The RGB-T comparison methods include QFDet~\citep{zhang2023rgbtdroneperson}, LF-DETR~\citep{qi2026lfdetr}, IM-CMDet~\citep{luo2025imcmdet}, COXNet~\citep{peng2025coxnet}, CCRDet~\citep{gu2026ccrdet}, C$^2$Former~\citep{yuan2024c2former}, TINet~\citep{tinet}, EI$^2$Det~\citep{hu2025ei2det}, and M$^2$D-LIF~\citep{Zhao_2025_ICCV}.
QFDet$^\ast$ and COXNet$^\ast$ denote the $l_2$--$l_6$ configurations of the corresponding methods. We use the published results for LF-DETR. We train the two dagger-marked models under the common RGB-T training protocol and reproduce the remaining methods using their original training settings.

\subsubsection{Metrics}
For RGBTDronePerson, we follow the benchmark evaluation protocol~\citep{zhang2023rgbtdroneperson} and report mAP$_{25}^{\mathrm{all}}$ and mAP$_{50}^{\mathrm{all}}$, which denote mean average precision at intersection over union (IoU) thresholds of 0.25 and 0.50, respectively. The superscript ``all'' indicates evaluation over all object sizes. We also report mAP$_{50}$ for the tiny, tiny1, tiny2, tiny3, and small subsets to assess detection performance at different scales. We measure object size in pixels as the square root of the ground-truth bounding-box area. The tiny1, tiny2, tiny3, and small subsets correspond to size ranges of $[0,8)$, $[8,12)$, $[12,20)$, and $[20,32]$, respectively. The tiny subset is the union of tiny1, tiny2, and tiny3.

For VTUAV-det and DroneVehicle, we adopt the COCO evaluation protocol~\citep{lin2014microsoft} and report mAP, mAP$_{50}$, mAP$_{75}$, mAP$_s$, mAP$_m$, and mAP$_l$. For the overall mAP, we compute average precision for each category at each IoU threshold from 0.50 to 0.95 in increments of 0.05, then take the mean of these values. The metrics mAP$_{50}$ and mAP$_{75}$ use IoU thresholds of 0.50 and 0.75, respectively. The scale-specific metrics mAP$_s$, mAP$_m$, and mAP$_l$ use the full range of IoU thresholds for small, medium, and large objects, respectively, with COCO area thresholds of $32^2$ and $96^2$ square pixels.

In addition, we report the number of parameters (M) and computational cost (GFLOPs) to characterize model complexity, and frames per second (FPS) to measure inference speed.

\subsubsection{Implementation Details}
MQCODet is built on DEIM~\citep{huang2025deim} with two independent HGNetv2-B0 backbones initialized from the same pretrained weights. The model is trained separately on each dataset using two NVIDIA RTX 5090 GPUs with a global batch size of 8 for up to 90 epochs with early stopping. We use AdamW with learning rates of $2.5\times10^{-5}$ for the backbone feature extractors and $5\times10^{-4}$ for the remaining parameters, a weight decay of $10^{-4}$, and a flat-cosine schedule. Training employs synchronized geometric augmentation, MixUp, CopyBlend, and multi-scale inputs of 480--800 pixels; evaluation uses $640\times640$ inputs. Parameters and GFLOPs are measured on the inference model using native MMCV 1.6.1 counting, treating one multiply--add as one operation. Inference speed is measured on an NVIDIA Jetson AGX Xavier using an optimized TensorRT FP16 engine with batch size 1, based on the combined latency of model inference and post-processing.

\subsection{Results on RGBTDronePerson}
\label{subsec:rgbtdroneperson_results}

\begin{table}[htbp]
\centering
\begin{minipage}{\linewidth}
\centering
\fontsize{9.5}{11.5}\selectfont
\setlength{\tabcolsep}{1pt}
\renewcommand{\arraystretch}{1.08}
\caption{Comparison with representative methods on RGBTDronePerson (\%). Bold values indicate the best results. $\dagger$ marks an RGB-T adaptation and $\ast$ denotes the $l_2$--$l_6$ configurations of QFDet and COXNet. LF-DETR results are taken from its original paper. Dashes indicate unreported values.}
\label{tab:main_comparison}
\begin{tabularx}{\linewidth}{@{}>{\raggedright\arraybackslash}p{0.35\linewidth}*{7}{Y}@{}}
\toprule
& \multicolumn{1}{c}{mAP$_{25}$} & \multicolumn{6}{c@{}}{mAP$_{50}$} \\
\cmidrule(lr){2-2}\cmidrule(l){3-8}
Method & all & tiny & tiny1 & tiny2 & tiny3 & small & all \\
\midrule
D-FINE$^{\dagger}$~\citep{peng2025dfine} & 58.5 & 48.3 & 15.6 & 35.5 & 53.0 & 27.9 & 47.6 \\
DEIM$^{\dagger}$~\citep{huang2025deim} & 59.7 & 49.8 & \textbf{29.1} & 39.6 & 54.6 & 28.3 & 48.9 \\
LF-DETR~\citep{qi2026lfdetr} & -- & 53.4 & -- & -- & -- & 30.2 & 50.9 \\
QFDet~\citep{zhang2023rgbtdroneperson} & 57.3 & 44.0 & 20.3 & 30.1 & 50.4 & 26.8 & 42.1 \\
QFDet$^{\ast}$~\citep{zhang2023rgbtdroneperson} & 61.6 & 48.8 & 22.2 & 37.9 & 53.7 & 28.4 & 46.7 \\
IM-CMDet~\citep{luo2025imcmdet} & 58.1 & 43.2 & 18.5 & 29.0 & 49.8 & 27.6 & 41.7 \\
COXNet~\citep{peng2025coxnet} & 59.0 & 47.2 & 27.4 & 35.6 & 52.6 & 29.7 & 45.6 \\
COXNet$^{\ast}$~\citep{peng2025coxnet} & 62.8 & 51.8 & 23.1 & 40.1 & 56.8 & 30.9 & 50.0 \\
CCRDet~\citep{gu2026ccrdet} & 57.9 & 44.2 & 17.8 & 32.2 & 49.9 & 29.0 & 42.3 \\
C$^2$Former~\citep{yuan2024c2former} & 42.8 & 44.6 & 7.1 & 34.9 & 49.1 & 18.2 & 34.9 \\
TINet~\citep{tinet} & 37.0 & 27.0 & 0.0 & 21.5 & 32.6 & 31.1 & 26.6 \\
EI$^2$Det~\citep{hu2025ei2det} & 43.6 & 44.5 & 7.7 & 30.3 & 53.8 & 26.5 & 36.0 \\
M$^2$D-LIF~\citep{Zhao_2025_ICCV} & 32.4 & 33.3 & 13.2 & 21.1 & 39.0 & 12.7 & 26.5 \\
\textbf{MQCODet (Ours)} & \textbf{64.1} & \textbf{55.7} & 26.4 & \textbf{46.0} & \textbf{58.9} & \textbf{32.0} & \textbf{54.7} \\
\bottomrule
\end{tabularx}
\end{minipage}
\end{table}

As shown in Table~\ref{tab:main_comparison}, MQCODet achieves the best results on six of the seven reported metrics, reaching 54.7\% mAP$_{50}^{\mathrm{all}}$ and 64.1\% mAP$_{25}^{\mathrm{all}}$. Compared with the DEIM$^{\dagger}$ baseline, MQCODet achieves gains of 5.8\% and 4.4\% in these two metrics, respectively. Its mAP$_{50}^{\mathrm{tiny}}$ also increases from 49.8\% to 55.7\%, demonstrating improved detection performance on tiny targets.

Compared with the representative RGB-T detectors, MQCODet surpasses LF-DETR by 3.8\% in mAP$_{50}^{\mathrm{all}}$ and 2.3\% in mAP$_{50}^{\mathrm{tiny}}$. It also outperforms COXNet$^{\ast}$ by 4.7\% and 1.3\% in mAP$_{50}^{\mathrm{all}}$ and mAP$_{25}^{\mathrm{all}}$, respectively. These comparisons show that MQCODet improves both overall detection accuracy and performance on the tiny subset.

The results across different size groups further characterize these gains. MQCODet achieves the highest mAP$_{50}$ on the tiny2, tiny3, and small subsets, reaching 46.0\%, 58.9\%, and 32.0\%, respectively. The largest improvement over DEIM$^{\dagger}$ occurs on tiny2, with a gain of 6.4\%; MQCODet also exceeds COXNet$^{\ast}$ by 5.9\% on this subset. However, its mAP$_{50}^{\mathrm{tiny1}}$ is 26.4\%, compared with 29.1\% for DEIM$^{\dagger}$. Detection of the smallest targets therefore remains an area for further improvement.

Fig.~\ref{fig:rgbtdrone_qualitative} further compares the detection results of CCRDet, QFDet, and MQCODet under low illumination, background clutter, and occlusion. MQCODet detects pedestrians missed by QFDet in the low-illumination scene and recovers some targets missed by CCRDet or QFDet in the cluttered and occluded scenes. These visual comparisons complement the quantitative results by showing fewer missed pedestrians in the illustrated scenes.

\begin{figure}[htbp]
\centering
\includegraphics[width=0.98\textwidth]{\manuscriptfigdir 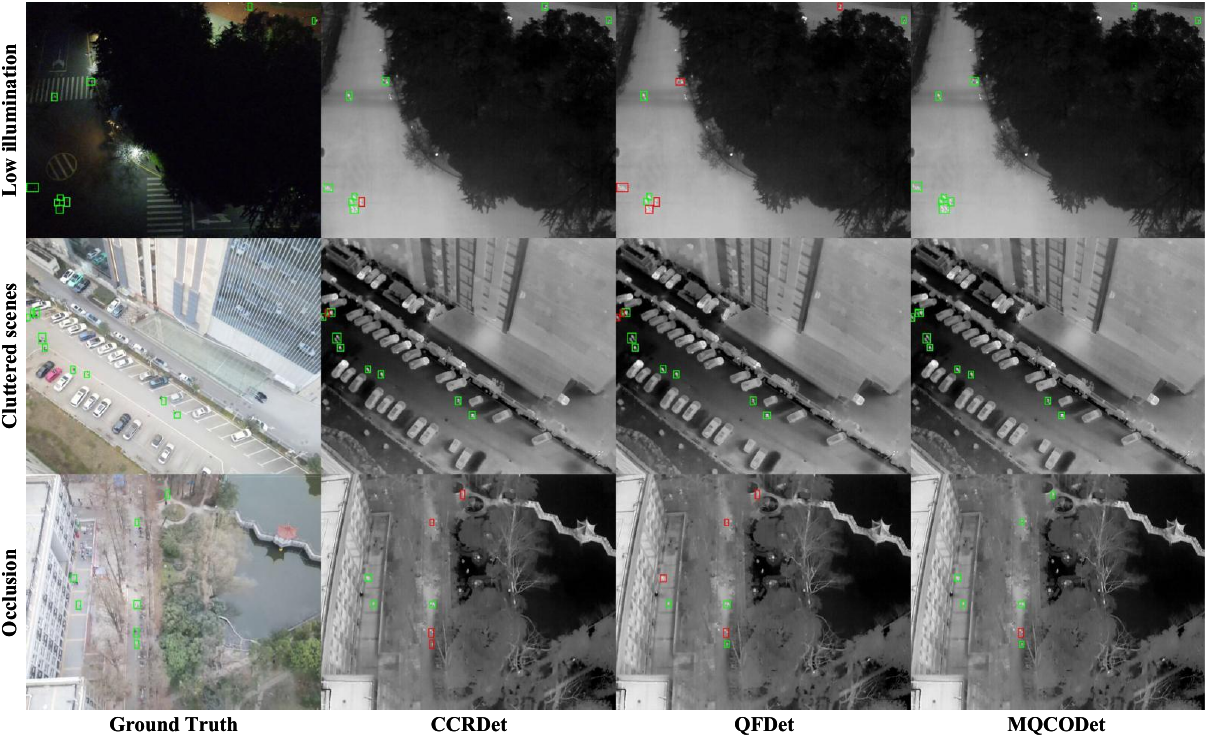}
\caption{Qualitative comparison on RGBTDronePerson under low illumination, background clutter, and occlusion (from top to bottom). The columns show the ground truth and the detection results of CCRDet, QFDet, and MQCODet, respectively. Green boxes mark ground-truth objects in the first column and correct detections in the remaining columns; red boxes indicate missed ground-truth objects.}
\label{fig:rgbtdrone_qualitative}
\end{figure}

\subsection{Results on VTUAV-det}
\label{subsec:vtuav_results}

\begin{table}[htbp]
\centering
\begin{minipage}{\linewidth}
\centering
\fontsize{9.5}{11.5}\selectfont
\setlength{\tabcolsep}{2pt}
\renewcommand{\arraystretch}{1.08}
\caption{Comparison with representative methods for person detection on VTUAV-det (\%). Bold values indicate the best results. $\dagger$ marks an RGB-T adaptation and $\ast$ denotes the $l_2$--$l_6$ configurations of QFDet and COXNet.}
\label{tab:vtuav_comparison}
\begin{tabularx}{\linewidth}{@{}>{\raggedright\arraybackslash}p{0.35\linewidth}*{6}{Y}@{}}
\toprule
Method & mAP & mAP$_{50}$ & mAP$_{75}$ & mAP$_s$ & mAP$_m$ & mAP$_l$ \\
\midrule
D-FINE$^{\dagger}$~\citep{peng2025dfine} & 32.8 & 74.6 & 25.4 & 21.0 & 30.9 & 56.1 \\
DEIM$^{\dagger}$~\citep{huang2025deim} & 33.0 & 73.7 & 26.1 & 21.8 & 30.8 & 57.4 \\
QFDet~\citep{zhang2023rgbtdroneperson} & 31.1 & 70.4 & 22.9 & 12.5 & 20.4 & 56.8 \\
QFDet$^{\ast}$~\citep{zhang2023rgbtdroneperson} & 33.3 & 75.5 & 24.2 & 18.1 & 32.4 & 57.2 \\
IM-CMDet~\citep{luo2025imcmdet} & 31.5 & 70.7 & 23.0 & 12.8 & 30.7 & 57.8 \\
COXNet~\citep{peng2025coxnet} & 31.5 & 71.8 & 23.1 & 15.3 & 30.6 & 56.0 \\
COXNet$^{\ast}$~\citep{peng2025coxnet} & 33.5 & 76.1 & 25.1 & 18.6 & 32.6 & 56.8 \\
CCRDet~\citep{gu2026ccrdet} & 32.6 & 73.1 & 24.4 & 14.8 & 32.1 & 57.6 \\
C$^2$Former~\citep{yuan2024c2former} & 28.2 & 64.5 & 20.6 & 11.2 & 27.8 & 53.2 \\
TINet~\citep{tinet} & 26.8 & 59.6 & 19.8 & 1.2 & 29.1 & 52.6 \\
EI$^2$Det~\citep{hu2025ei2det} & 32.9 & 76.1 & 25.3 & 17.1 & 33.2 & 57.9 \\
M$^2$D-LIF~\citep{Zhao_2025_ICCV} & 32.1 & 68.9 & 24.9 & 13.4 & 31.6 & 57.5 \\
\textbf{MQCODet (Ours)} & \textbf{37.1} & \textbf{79.4} & \textbf{29.5} & \textbf{24.1} & \textbf{35.2} & \textbf{60.4} \\
\bottomrule
\end{tabularx}
\end{minipage}
\end{table}

\begin{figure}[htbp]
\centering
\includegraphics[width=0.98\textwidth]{\manuscriptfigdir 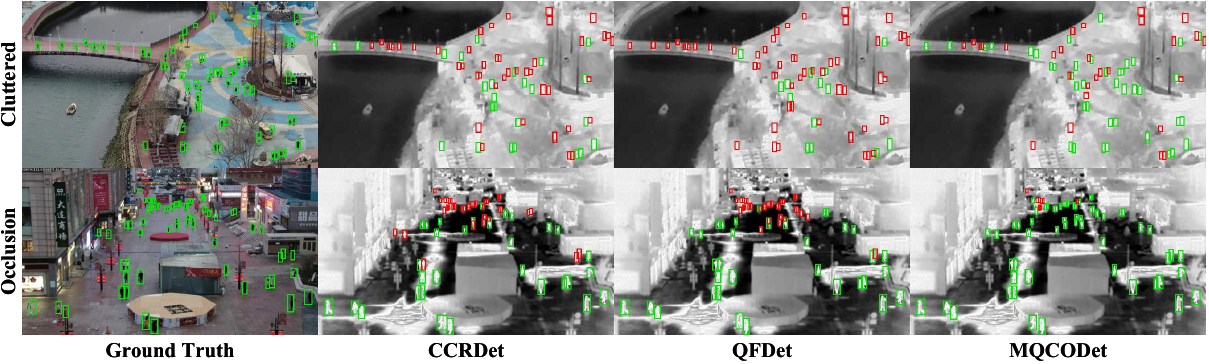}
\caption{Qualitative comparison for person detection on VTUAV-det under background clutter and partial occlusion. Green boxes mark ground-truth objects in the first column and correct detections in the remaining columns. Red boxes indicate missed ground-truth objects.}
\label{fig:vtuav_qualitative}
\end{figure}

As shown in Table~\ref{tab:vtuav_comparison}, MQCODet achieves the best results across all six metrics, obtaining 37.1\% mAP and 79.4\% mAP$_{50}$. Compared with the DEIM$^{\dagger}$ baseline, MQCODet achieves absolute gains of 4.1\% and 5.7\% in these two metrics, respectively. It also outperforms COXNet$^{\ast}$ by 3.6\% and 3.3\% in mAP and mAP$_{50}$, respectively.
Under the stricter IoU threshold of 0.75, MQCODet achieves 29.5\% mAP$_{75}$, surpassing DEIM$^{\dagger}$ by 3.4\%. Furthermore, MQCODet obtains 24.1\%, 35.2\%, and 60.4\% mAP on small, medium, and large pedestrians, respectively, improving over DEIM$^{\dagger}$ by 2.3\%, 4.4\%, and 3.0\%. These results demonstrate improved detection performance under stricter localization requirements and across different pedestrian scales.
Fig.~\ref{fig:vtuav_qualitative} further presents qualitative comparisons of person detection in VTUAV-det scenes with cluttered backgrounds and partial occlusion.

\subsection{Results on DroneVehicle}
\label{subsec:dronevehicle_results}

\begin{table}[htbp]
\centering
\begin{minipage}{\linewidth}
\centering
\fontsize{9.5}{11.5}\selectfont
\setlength{\tabcolsep}{2pt}
\renewcommand{\arraystretch}{1.08}
\caption{Comparison with representative methods for vehicle detection on the DroneVehicle validation set (\%). Bold values indicate the best results. $\dagger$ marks an RGB-T adaptation and $\ast$ denotes the $l_2$--$l_6$ configurations of QFDet and COXNet.}
\label{tab:dronevehicle_comparison}
\begin{tabularx}{\linewidth}{@{}>{\raggedright\arraybackslash}p{0.35\linewidth}*{6}{Y}@{}}
\toprule
Method & mAP & mAP$_{50}$ & mAP$_{75}$ & mAP$_s$ & mAP$_m$ & mAP$_l$ \\
\midrule
D-FINE$^{\dagger}$~\citep{peng2025dfine} & 61.7 & 81.6 & 73.7 & 31.5 & 61.6 & 80.6 \\
DEIM$^{\dagger}$~\citep{huang2025deim} & 63.4 & 83.9 & 74.8 & 32.2 & 63.4 & 81.7 \\
QFDet~\citep{zhang2023rgbtdroneperson} & 55.5 & 78.1 & 67.1 & 20.6 & 56.1 & 73.1 \\
QFDet$^{\ast}$~\citep{zhang2023rgbtdroneperson} & 56.4 & 79.6 & 68.0 & 21.2 & 57.0 & 73.6 \\
COXNet~\citep{peng2025coxnet} & 53.5 & 75.2 & 64.3 & 18.5 & 55.2 & 68.4 \\
COXNet$^{\ast}$~\citep{peng2025coxnet} & 53.9 & 75.8 & 65.4 & 21.2 & 55.5 & 68.5 \\
CCRDet~\citep{gu2026ccrdet} & 55.7 & 78.5 & 67.2 & 23.5 & 56.4 & 72.9 \\
C$^2$Former~\citep{yuan2024c2former} & 45.7 & 71.3 & 53.4 & 19.6 & 47.9 & 43.7 \\
TINet~\citep{tinet} & 54.1 & 77.4 & 65.3 & 18.4 & 54.6 & 70.3 \\
EI$^2$Det~\citep{hu2025ei2det} & 61.2 & 83.5 & 73.6 & 26.6 & 61.6 & 77.4 \\
M$^2$D-LIF~\citep{Zhao_2025_ICCV} & 59.9 & 79.2 & 71.6 & 28.1 & 59.9 & 78.8 \\
\textbf{MQCODet (Ours)} & \textbf{66.1} & \textbf{86.3} & \textbf{78.0} & \textbf{32.8} & \textbf{66.0} & \textbf{84.2} \\
\bottomrule
\end{tabularx}
\end{minipage}
\end{table}

As shown in Table~\ref{tab:dronevehicle_comparison}, MQCODet achieves the best results across all six metrics on the DroneVehicle validation set, obtaining 66.1\% mAP, 86.3\% mAP$_{50}$, and 78.0\% mAP$_{75}$. Compared with the DEIM$^{\dagger}$ baseline, these results represent absolute gains of 2.7\%, 2.4\%, and 3.2\%, respectively. MQCODet also outperforms EI$^2$Det by 4.9\%, 2.8\%, and 4.4\% in these three metrics, respectively.
Across object scales, MQCODet achieves 32.8\% mAP$_s$, 66.0\% mAP$_m$, and 84.2\% mAP$_l$, improving over DEIM$^{\dagger}$ by 0.6\%, 2.6\%, and 2.5\%, respectively. The gains are larger for medium and large vehicles, while a smaller improvement is observed for small vehicles. These results support the applicability of MQCODet to vehicle detection across different object scales.
Fig.~\ref{fig:dronevehicle_qualitative} further presents qualitative comparisons of vehicle detection under low illumination and in cluttered road scenes.

\begin{figure}[htbp]
\centering
\includegraphics[width=0.98\textwidth]{\manuscriptfigdir 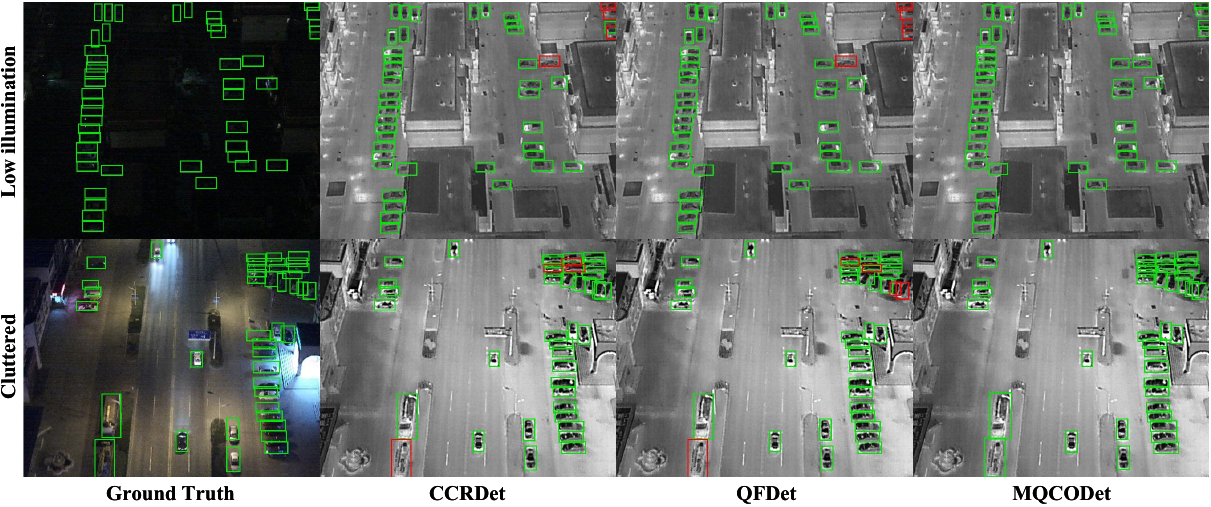}
\caption{Qualitative comparison for vehicle detection on DroneVehicle under low illumination and in cluttered road scenes. Green boxes mark ground-truth objects in the first column and correct detections in the remaining columns. Red boxes indicate missed ground-truth objects.}
\label{fig:dronevehicle_qualitative}
\end{figure}

\subsection{Ablation Studies}
\label{subsec:ablations}
We conduct ablation studies on RGBTDronePerson to evaluate the contributions of the proposed components and examine the design choices of DCRF, ECI, and MAA.

\subsubsection{Contribution of the Main Components}

All configurations in Table~\ref{tab:ablation_components} retain the dual-branch HGNetv2-B0 backbone architecture of DEIM$^\dagger$. When DCRF is disabled, we average the RGB and thermal feature maps element-wise at each level. Enabling DCRF replaces this average with alignment and region-aware fusion.

\begin{table}[htbp]
\centering
\begin{minipage}{\linewidth}
\centering
\fontsize{9.5}{11.5}\selectfont
\setlength{\tabcolsep}{2pt}
\renewcommand{\arraystretch}{1.08}
\caption{Ablation of DCRF, ECI, and MAA on RGBTDronePerson. The first row is the dual-branch DEIM$^{\dagger}$ baseline.}
\label{tab:ablation_components}
\begin{tabularx}{\linewidth}{@{}>{\raggedright\arraybackslash}p{0.15\linewidth}*{6}{Y}@{}}
\toprule
DCRF & ECI & MAA & mAP$_{50}^{\mathrm{all}}$ & mAP$_{25}^{\mathrm{all}}$ & GFLOPs & Params (M) \\
\midrule
-- & -- & -- & 48.9 & 59.7 & 21.1 & 9.2 \\
$\checkmark$ & -- & -- & 51.3 & 61.8 & 19.6 & 9.0 \\
-- & $\checkmark$ & -- & 50.4 & 61.3 & 21.1 & 9.4 \\
$\checkmark$ & $\checkmark$ & -- & 52.5 & 63.4 & 19.6 & 9.1 \\
$\checkmark$ & -- & $\checkmark$ & 53.2 & 63.6 & 19.6 & 9.0 \\
$\checkmark$ & $\checkmark$ & $\checkmark$ & 54.7 & 64.1 & 19.6 & 9.1 \\
\bottomrule
\end{tabularx}
\end{minipage}
\end{table}

Adding DCRF and ECI individually to the DEIM$^{\dagger}$ baseline yields absolute gains of 2.4\% and 1.5\% in mAP$_{50}^{\mathrm{all}}$, respectively. Combining DCRF and ECI raises this metric to 52.5\%, improving over the DCRF configuration by 1.2\%. MAA further improves mAP$_{50}^{\mathrm{all}}$ by 1.9\% when used with DCRF and by 2.2\% when used with both DCRF and ECI. The complete model achieves 54.7\% mAP$_{50}^{\mathrm{all}}$ and 64.1\% mAP$_{25}^{\mathrm{all}}$, exceeding the baseline by 5.8\% and 4.4\%, respectively.

The complete model uses 9.1 M parameters and 19.6 GFLOPs, compared with 9.2 M parameters and 21.1 GFLOPs for the baseline. Enabling MAA in the DCRF--ECI configuration improves accuracy while leaving the reported parameter count and GFLOPs unchanged. Fig.~\ref{fig:ablation_heatmap} further presents a visual comparison of the fused feature responses for the baseline, the configuration with DCRF, and the complete model.

\begin{figure}[htbp]
\centering
\includegraphics[width=\columnwidth]{\manuscriptfigdir 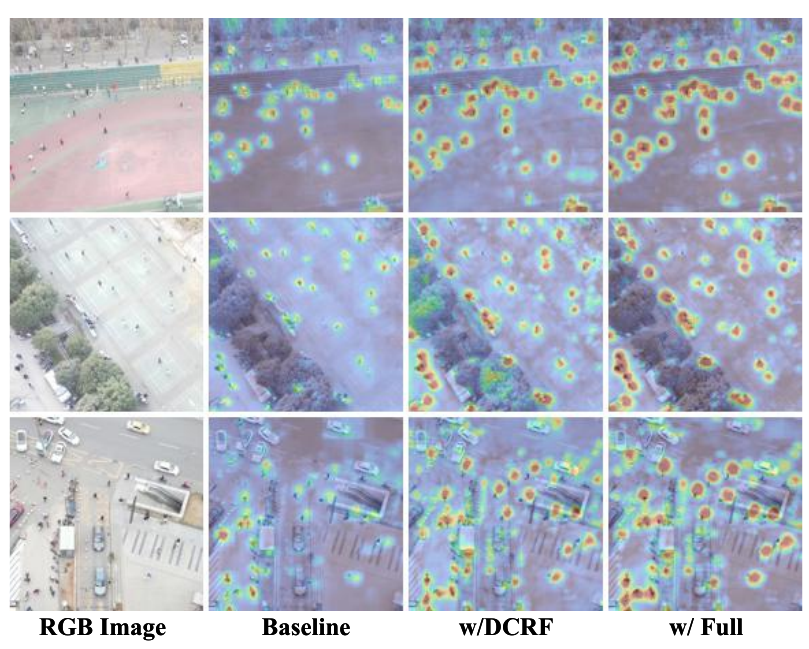}
\caption{Modality-fused feature responses on RGBTDronePerson for the baseline, the configuration with DCRF, and the complete model.}
\label{fig:ablation_heatmap}
\end{figure}

\subsubsection{DCRF Component Analysis}

We first examine the roles of modality-aware fusion, feature alignment, and the scene-level quality prior in Table~\ref{tab:ablation_dcrf}. Within this comparison, modality-aware fusion without alignment or the quality prior achieves 52.2\% mAP$_{50}^{\mathrm{all}}$ and 62.6\% mAP$_{25}^{\mathrm{all}}$. Introducing feature alignment raises these scores to 53.5\% and 63.4\%, gains of 1.3 and 0.8 percentage points, respectively. Adding the quality prior raises them to 54.7\% and 64.1\%, a further 1.2 and 0.7 percentage points, respectively. These results support the combined use of alignment and quality information. Fig.~\ref{fig:dcrf_alignment_visualization} provides a visual comparison of feature responses without and with DCRF alignment.

\begin{table}[htbp]
\centering
\begin{minipage}{\linewidth}
\centering
\fontsize{9.5}{11.5}\selectfont
\setlength{\tabcolsep}{2pt}
\renewcommand{\arraystretch}{1.08}
\caption{Ablation of region-aware modality fusion, feature alignment, and the scene-level quality prior on RGBTDronePerson.}
\label{tab:ablation_dcrf}
\begin{tabularx}{\linewidth}{@{}>{\raggedright\arraybackslash}p{0.15\linewidth}*{4}{Y}@{}}
\toprule
Modality & Align & Prior & mAP$_{50}^{\mathrm{all}}$ & mAP$_{25}^{\mathrm{all}}$ \\
\midrule
$\checkmark$ & -- & -- & 52.2 & 62.6 \\
$\checkmark$ & $\checkmark$ & -- & 53.5 & 63.4 \\
$\checkmark$ & $\checkmark$ & $\checkmark$ & 54.7 & 64.1 \\
\bottomrule
\end{tabularx}
\end{minipage}
\end{table}

\begin{figure}[htbp]
\centering
\includegraphics[width=\columnwidth]{\manuscriptfigdir 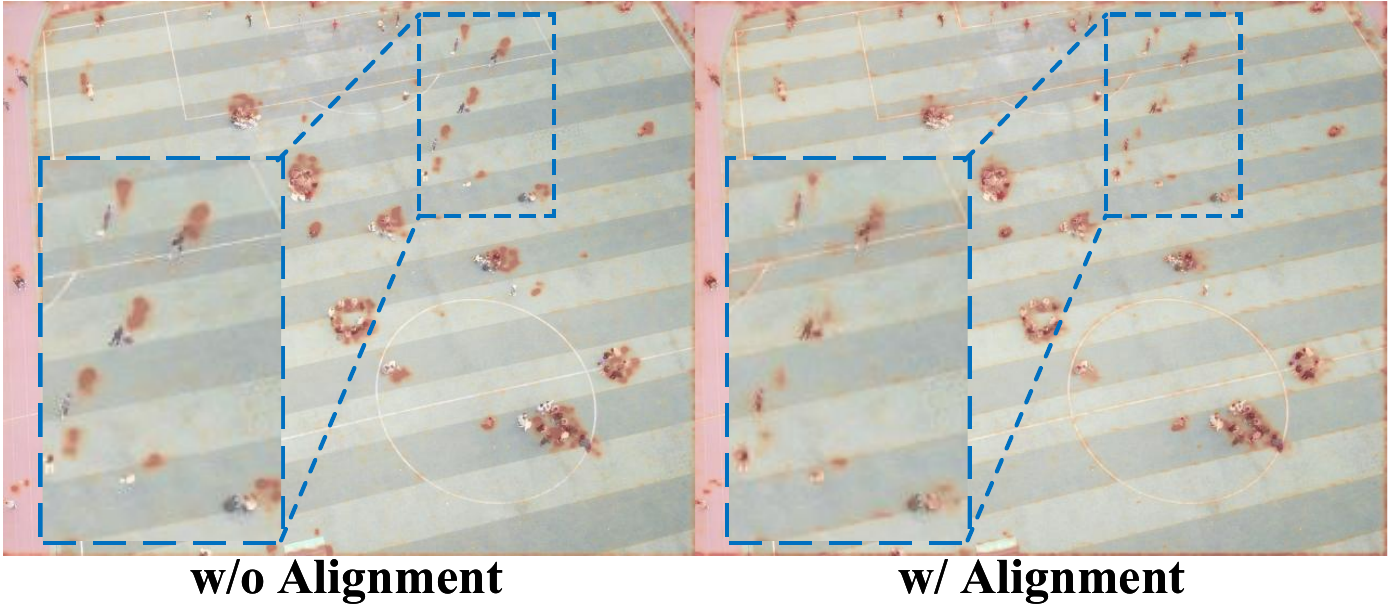}
\caption{Feature response comparison without alignment and with DCRF alignment. Orange regions indicate strong responses.}
\label{fig:dcrf_alignment_visualization}
\end{figure}

\begin{table}[htbp]
\centering
\begin{minipage}{\linewidth}
\centering
\fontsize{9.5}{11.5}\selectfont
\setlength{\tabcolsep}{2pt}
\renewcommand{\arraystretch}{1.08}
\caption{Comparison of cross-modal alignment directions on RGBTDronePerson.}
\label{tab:ablation_dcrf_alignment_direction}
\begin{tabularx}{\linewidth}{@{}>{\raggedright\arraybackslash}p{0.15\linewidth}*{7}{Y}@{}}
\toprule
Direction
& mAP$_{25}^{\mathrm{all}}$
& mAP$_{50}^{\mathrm{tiny}}$
& mAP$_{50}^{\mathrm{tiny1}}$
& mAP$_{50}^{\mathrm{tiny2}}$
& mAP$_{50}^{\mathrm{tiny3}}$
& mAP$_{50}^{\mathrm{small}}$
& mAP$_{50}^{\mathrm{all}}$ \\
\midrule
RGB $\rightarrow$ T & 64.1 & 55.7 & 26.4 & 46.0 & 58.9 & 32.0 & 54.7 \\
T $\rightarrow$ RGB & 63.8 & 55.0 & 23.9 & 42.8 & 60.2 & 32.7 & 53.9 \\
Bidirectional & 63.3 & 54.1 & 22.8 & 42.2 & 58.6 & 30.8 & 53.0 \\
\bottomrule
\end{tabularx}
\end{minipage}
\end{table}

To determine the alignment direction, Table~\ref{tab:ablation_dcrf_alignment_direction} compares RGB-to-thermal, thermal-to-RGB, and bidirectional alignment. RGB-to-thermal alignment achieves the highest mAP$_{50}^{\mathrm{all}}$ of 54.7\%, exceeding the reverse and bidirectional configurations by 0.8\% and 1.7\%, respectively. It also obtains the best results on the tiny1 and tiny2 subsets, although thermal-to-RGB alignment performs better on tiny3 and small objects. We therefore adopt RGB-to-thermal alignment based on its overall detection performance.

\begin{table}[htbp]
\centering
\begin{minipage}{\linewidth}
\centering
\fontsize{9.5}{11.5}\selectfont
\setlength{\tabcolsep}{2pt}
\renewcommand{\arraystretch}{1.08}
\caption{Effect of the offset bound $s_{\Delta}$ on RGBTDronePerson. The bound is measured in feature-grid cells.}
\label{tab:ablation_offset}
\begin{tabularx}{\linewidth}{@{}>{\raggedright\arraybackslash}p{0.15\linewidth}*{7}{Y}@{}}
\toprule
$s_{\Delta}$
& mAP$_{25}^{\mathrm{all}}$
& mAP$_{50}^{\mathrm{tiny}}$
& mAP$_{50}^{\mathrm{tiny1}}$
& mAP$_{50}^{\mathrm{tiny2}}$
& mAP$_{50}^{\mathrm{tiny3}}$
& mAP$_{50}^{\mathrm{small}}$
& mAP$_{50}^{\mathrm{all}}$ \\
\midrule
0 & 63.8 & 54.5 & 24.0 & 43.0 & 58.2 & 30.6 & 53.2 \\
1 & 64.2 & 55.1 & 25.1 & 44.2 & 59.1 & 31.5 & 53.8 \\
2 & 64.1 & 55.7 & 26.4 & 46.0 & 58.9 & 32.0 & 54.7 \\
4 & 63.6 & 54.9 & 25.7 & 42.7 & 58.5 & 31.2 & 53.7 \\
\bottomrule
\end{tabularx}
\end{minipage}
\end{table}

The effect of the offset bound $s_{\Delta}$ is evaluated in Table~\ref{tab:ablation_offset}. Increasing the bound from 0 to 2 improves mAP$_{50}^{\mathrm{all}}$ from 53.2\% to 54.7\%, whereas further increasing it to 4 reduces the result to 53.7\%. The setting $s_{\Delta}=2$ also achieves the highest mAP$_{50}$ on the tiny, tiny1, tiny2, and small subsets. We select this bound because it gives the highest mAP$_{50}^{\mathrm{all}}$ among the evaluated settings.

\subsubsection{Multi-Scale Encoder Comparison}

\begin{table}[htbp]
\centering
\begin{minipage}{\linewidth}
\centering
\fontsize{9.5}{11.5}\selectfont
\setlength{\tabcolsep}{2pt}
\renewcommand{\arraystretch}{1.08}
\caption{Comparison of multi-scale encoders within the MQCODet framework on RGBTDronePerson.}
\label{tab:ablation_eci_comparison}
\begin{tabularx}{\linewidth}{@{}>{\raggedright\arraybackslash}p{0.43\linewidth}*{4}{Y}@{}}
\toprule
Encoder & mAP$_{50}^{\mathrm{all}}$ & mAP$_{25}^{\mathrm{all}}$ & Params (M) & GFLOPs \\
\midrule
BiFPN~\citep{tan2020efficientdet} & 53.0 & 63.2 & 7.9 & 14.0 \\
GDecoder~\citep{wang2023gold} & 55.4 & 64.9 & 13.6 & 42.7 \\
HyperEncoder~\citep{peng2025dfine} & 53.4 & 63.6 & 9.7 & 24.8 \\
HyperEncoderV2~\citep{huang2025deimv2} & 53.5 & 64.0 & 9.0 & 19.6 \\
ECI (Ours) & 54.7 & 64.1 & 9.1 & 19.6 \\
\bottomrule
\end{tabularx}
\end{minipage}
\end{table}

Table~\ref{tab:ablation_eci_comparison} compares ECI with alternative multi-scale encoders, including BiFPN~\citep{tan2020efficientdet} and GDecoder~\citep{wang2023gold}. ECI achieves 54.7\% mAP$_{50}^{\mathrm{all}}$ with 9.1 M parameters and 19.6 GFLOPs. Compared with HyperEncoderV2, ECI improves this metric by 1.2\% at the same reported GFLOPs, with only 0.1 M additional parameters. GDecoder achieves a higher mAP$_{50}^{\mathrm{all}}$ of 55.4\%, but requires 13.6 M parameters and 42.7 GFLOPs. ECI reduces these costs by approximately 33.1\% and 54.1\%, respectively, with mAP$_{50}^{\mathrm{all}}$ lower by 0.7\%. These comparisons motivate the use of ECI to balance detection accuracy and computational cost.

\subsubsection{MAA Loss and Parameter Analysis}

\begin{table}[htbp]
\centering
\begin{minipage}{\linewidth}
\centering
\fontsize{9.5}{11.5}\selectfont
\setlength{\tabcolsep}{2pt}
\renewcommand{\arraystretch}{1.08}
\caption{Ablation of the localization terms, scene-level quality prior, and instance-rank residual in MAA on RGBTDronePerson.}
\label{tab:ablation_maa}
\begin{tabularx}{\linewidth}{@{}>{\raggedright\arraybackslash}p{0.15\linewidth}*{5}{Y}@{}}
\toprule
NWD & GIoU & Image prior & Rank residual & mAP$_{50}^{\mathrm{all}}$ & mAP$_{25}^{\mathrm{all}}$ \\
\midrule
$\checkmark$ & -- & -- & -- & 52.7 & 63.2 \\
-- & $\checkmark$ & -- & -- & 52.5 & 63.4 \\
$\checkmark$ & $\checkmark$ & $\checkmark$ & -- & 53.7 & 63.6 \\
$\checkmark$ & $\checkmark$ & -- & $\checkmark$ & 53.5 & 63.4 \\
$\checkmark$ & $\checkmark$ & $\checkmark$ & $\checkmark$ & 54.7 & 64.1 \\
\bottomrule
\end{tabularx}
\end{minipage}
\end{table}

We evaluate the localization terms and the two modulation components of MAA in Table~\ref{tab:ablation_maa}. The NWD and GIoU configurations achieve 52.7\% and 52.5\% mAP$_{50}^{\mathrm{all}}$, respectively. With both localization terms active, the image prior alone ($\eta=0$) yields 53.7\% mAP$_{50}^{\mathrm{all}}$ and 63.6\% mAP$_{25}^{\mathrm{all}}$. The rank residual alone ($\beta_g=0$) yields 53.5\% and 63.4\%, respectively. With both components, the scores reach 54.7\% and 64.1\%. Adding the rank residual to the image-prior configuration improves the two metrics by 1.0 and 0.5 percentage points, respectively. Adding the image prior to the rank-residual configuration improves them by 1.2 and 0.7 percentage points, respectively. These comparisons indicate that the two forms of modulation provide complementary benefits in the evaluated configurations.

\begin{table}[htbp]
\centering
\begin{minipage}{\linewidth}
\centering
\fontsize{9.5}{11.5}\selectfont
\setlength{\tabcolsep}{2pt}
\renewcommand{\arraystretch}{1.08}
\caption{Effect of the base temperature $\alpha_g$ on RGBTDronePerson.}
\label{tab:maa_alpha}
\begin{tabularx}{\linewidth}{@{}>{\raggedright\arraybackslash}p{0.15\linewidth}*{7}{Y}@{}}
\toprule
$\alpha_g$
& mAP$_{25}^{\mathrm{all}}$
& mAP$_{50}^{\mathrm{tiny}}$
& mAP$_{50}^{\mathrm{tiny1}}$
& mAP$_{50}^{\mathrm{tiny2}}$
& mAP$_{50}^{\mathrm{tiny3}}$
& mAP$_{50}^{\mathrm{small}}$
& mAP$_{50}^{\mathrm{all}}$ \\
\midrule
0.5 & 63.8 & 54.7 & 20.7 & 43.4 & 59.8 & 32.3 & 54.5 \\
1.0 & 64.1 & 55.7 & 26.4 & 46.0 & 58.9 & 32.0 & 54.7 \\
1.5 & 64.3 & 54.8 & 26.8 & 42.4 & 58.8 & 32.5 & 52.8 \\
2.0 & 63.4 & 54.0 & 26.8 & 42.8 & 59.1 & 33.3 & 53.1 \\
\bottomrule
\end{tabularx}
\end{minipage}
\end{table}

For the base temperature $\alpha_g$, Table~\ref{tab:maa_alpha} shows that $\alpha_g=1.0$ achieves the highest mAP$_{50}^{\mathrm{all}}$ of 54.7\%. Compared with $\alpha_g=0.5$, this setting improves mAP$_{50}^{\mathrm{tiny1}}$ and mAP$_{50}^{\mathrm{tiny2}}$ by 5.7\% and 2.6\%, respectively. Increasing $\alpha_g$ to 1.5 or 2.0 reduces mAP$_{50}^{\mathrm{all}}$ to 52.8\% and 53.1\%.

\begin{table}[htbp]
\centering
\begin{minipage}{\linewidth}
\centering
\fontsize{9.5}{11.5}\selectfont
\setlength{\tabcolsep}{2pt}
\renewcommand{\arraystretch}{1.08}
\caption{Effect of the quality-temperature coefficient $\beta_g$ on RGBTDronePerson.}
\label{tab:maa_beta}
\begin{tabularx}{\linewidth}{@{}>{\raggedright\arraybackslash}p{0.15\linewidth}*{7}{Y}@{}}
\toprule
$\beta_g$
& mAP$_{25}^{\mathrm{all}}$
& mAP$_{50}^{\mathrm{tiny}}$
& mAP$_{50}^{\mathrm{tiny1}}$
& mAP$_{50}^{\mathrm{tiny2}}$
& mAP$_{50}^{\mathrm{tiny3}}$
& mAP$_{50}^{\mathrm{small}}$
& mAP$_{50}^{\mathrm{all}}$ \\
\midrule
0.0 & 63.4 & 54.9 & 26.7 & 43.4 & 60.1 & 30.6 & 53.5 \\
0.5 & 64.0 & 53.6 & 27.2 & 42.2 & 58.7 & 31.5 & 52.5 \\
1.0 & 64.1 & 55.7 & 26.4 & 46.0 & 58.9 & 32.0 & 54.7 \\
1.5 & 63.8 & 54.1 & 26.2 & 42.5 & 58.4 & 32.3 & 53.2 \\
\bottomrule
\end{tabularx}
\end{minipage}
\end{table}

Table~\ref{tab:maa_beta} examines the quality-temperature coefficient $\beta_g$. The setting $\beta_g=1.0$ achieves 54.7\% mAP$_{50}^{\mathrm{all}}$ and 64.1\% mAP$_{25}^{\mathrm{all}}$, improving over $\beta_g=0$ by 1.2 and 0.7 percentage points, respectively. Both $\beta_g=0.5$ and $\beta_g=1.5$ yield lower mAP$_{50}^{\mathrm{all}}$, supporting the choice of $\beta_g=1.0$ for quality-dependent modulation of the base gate.

\begin{table}[htbp]
\centering
\begin{minipage}{\linewidth}
\centering
\fontsize{9.5}{11.5}\selectfont
\setlength{\tabcolsep}{2pt}
\renewcommand{\arraystretch}{1.08}
\caption{Effect of the instance-rank residual amplitude $\eta$ on RGBTDronePerson.}
\label{tab:maa_eta}
\begin{tabularx}{\linewidth}{@{}>{\raggedright\arraybackslash}p{0.15\linewidth}*{7}{Y}@{}}
\toprule
$\eta$
& mAP$_{25}^{\mathrm{all}}$
& mAP$_{50}^{\mathrm{tiny}}$
& mAP$_{50}^{\mathrm{tiny1}}$
& mAP$_{50}^{\mathrm{tiny2}}$
& mAP$_{50}^{\mathrm{tiny3}}$
& mAP$_{50}^{\mathrm{small}}$
& mAP$_{50}^{\mathrm{all}}$ \\
\midrule
0 & 63.6 & 54.5 & 23.6 & 42.5 & 60.4 & 32.2 & 53.7 \\
0.01 & 63.8 & 54.8 & 23.4 & 42.2 & 60.5 & 30.9 & 53.6 \\
0.02 & 64.1 & 55.7 & 26.4 & 46.0 & 58.9 & 32.0 & 54.7 \\
0.03 & 64.3 & 54.0 & 24.6 & 41.7 & 59.3 & 31.5 & 52.8 \\
\bottomrule
\end{tabularx}
\end{minipage}
\end{table}

The effect of the instance-rank residual amplitude $\eta$ is shown in Table~\ref{tab:maa_eta}. Setting $\eta=0.02$ yields the highest mAP$_{50}^{\mathrm{all}}$ of 54.7\%. Relative to $\eta=0$, mAP$_{50}^{\mathrm{all}}$ and mAP$_{25}^{\mathrm{all}}$ increase by 1.0 and 0.5 percentage points, respectively. The same change raises mAP$_{50}^{\mathrm{tiny1}}$ and mAP$_{50}^{\mathrm{tiny2}}$ by 2.8 and 3.5 percentage points, respectively. Increasing the amplitude to 0.03 reduces mAP$_{50}^{\mathrm{all}}$ to 52.8\%, indicating that a larger residual amplitude does not necessarily improve detection performance.

\begin{table}[htbp]
\centering
\begin{minipage}{\linewidth}
\centering
\fontsize{9.5}{11.5}\selectfont
\setlength{\tabcolsep}{2pt}
\renewcommand{\arraystretch}{1.08}
\caption{Effect of the scale parameter $C_g$ on RGBTDronePerson.}
\label{tab:maa_scale}
\begin{tabularx}{\linewidth}{@{}>{\raggedright\arraybackslash}p{0.15\linewidth}*{7}{Y}@{}}
\toprule
$C_g$
& mAP$_{25}^{\mathrm{all}}$
& mAP$_{50}^{\mathrm{tiny}}$
& mAP$_{50}^{\mathrm{tiny1}}$
& mAP$_{50}^{\mathrm{tiny2}}$
& mAP$_{50}^{\mathrm{tiny3}}$
& mAP$_{50}^{\mathrm{small}}$
& mAP$_{50}^{\mathrm{all}}$ \\
\midrule
8 & 63.5 & 54.4 & 26.7 & 42.2 & 59.9 & 31.2 & 53.1 \\
16 & 63.8 & 54.5 & 26.3 & 42.3 & 60.0 & 32.9 & 53.6 \\
32 & 64.1 & 55.7 & 26.4 & 46.0 & 58.9 & 32.0 & 54.7 \\
48 & 64.0 & 54.2 & 25.8 & 42.6 & 59.4 & 32.5 & 53.3 \\
\bottomrule
\end{tabularx}
\end{minipage}
\end{table}

Finally, Table~\ref{tab:maa_scale} shows that $C_g=32$ achieves the highest mAP$_{50}^{\mathrm{all}}$ of 54.7\%, exceeding the results at $C_g=8$, 16, and 48 by 1.6\%, 1.1\%, and 1.4\%, respectively. This setting also obtains the highest mAP$_{50}^{\mathrm{tiny}}$ and mAP$_{50}^{\mathrm{tiny2}}$, reaching 55.7\% and 46.0\%, although $C_g=16$ gives a higher result on the small subset. Based on mAP$_{50}^{\mathrm{all}}$ in these comparisons, we use $\alpha_g=1.0$, $\beta_g=1.0$, $\eta=0.02$, and $C_g=32$ in the final configuration.

\subsection{Deployment on an Embedded Platform}
\label{subsec:embedded_deployment}

We benchmark MQCODet and the comparison methods on an NVIDIA Jetson AGX Xavier using TensorRT 7.1.0.16 with FP16 precision. Each run uses a batch size of 1 and $640\times640$ RGB-T image pairs. The latency in Table~\ref{tab:embedded_deployment} includes model inference and post-processing. We compute FPS from this combined latency.

\begin{table}[htbp]
\centering
\begin{minipage}{\linewidth}
\centering
\fontsize{9.5}{11.5}\selectfont
\setlength{\tabcolsep}{5pt}
\renewcommand{\arraystretch}{1.12}
\caption{Model complexity, detection accuracy (\%) on RGBTDronePerson, and embedded inference performance on the NVIDIA Jetson AGX Xavier. FPS and latency include model inference and post-processing with TensorRT FP16. Bold and underlined values indicate the best and second-best results in each column, respectively.}
\label{tab:embedded_deployment}
\begin{tabular}{lccccc}
\toprule
Method & Params (M) & GFLOPs & mAP$_{50}^{\mathrm{all}}$ & FPS$\uparrow$ & Latency (ms)$\downarrow$ \\
\midrule
QFDet~\citep{zhang2023rgbtdroneperson} & 60.6 & 203.6 & 42.1 & 6.3 & 158.3 \\
CCRDet~\citep{gu2026ccrdet} & 71.8 & 154.3 & 42.3 & 5.3 & 190.2 \\
DEIM$^{\dagger}$~\citep{huang2025deim} & \underline{9.2} & \underline{21.1} & \underline{48.9} & \textbf{32.8} & \textbf{30.5} \\
MQCODet & \textbf{9.1} & \textbf{19.6} & \textbf{54.7} & \underline{30.6} & \underline{32.7} \\
\bottomrule
\end{tabular}
\end{minipage}
\end{table}

As shown in Table~\ref{tab:embedded_deployment}, MQCODet achieves an inference speed of 30.6 FPS with a latency of 32.7 ms on the NVIDIA Jetson AGX Xavier, demonstrating its favorable deployment efficiency. Compared with CCRDet and QFDet, MQCODet achieves approximately 5.8 and 4.9 times their respective throughput, while requiring considerably fewer parameters and lower computational costs. In terms of detection accuracy, MQCODet achieves 54.7\% mAP$_{50}^{\mathrm{all}}$, yielding absolute gains of 12.4\% and 12.6\% over CCRDet and QFDet, respectively. These results demonstrate the advantages of MQCODet in both detection accuracy and computational efficiency.

Compared with DEIM$^{\dagger}$, MQCODet improves mAP$_{50}^{\mathrm{all}}$ from 48.9\% to 54.7\%, achieving an absolute gain of 5.8\% while maintaining a comparable inference speed (30.6 FPS vs. 32.8 FPS). Moreover, MQCODet reduces the parameter count and computational complexity. These results demonstrate that MQCODet achieves a favorable trade-off between detection accuracy and inference efficiency, supporting its practical deployment on resource-constrained embedded platforms.

\endgroup

\section{Conclusion}
\label{sec:conclusion}
In this paper, we proposed a Modality Quality-Guided Collaborative Optimization Detector (MQCODet) to address modality quality imbalance and spatial position misalignment in UAV-based RGB-T tiny pedestrian detection. Specifically, the DCRF module performs cross-modal feature calibration and region-aware fusion, while the ECI module enhances multi-scale contextual interaction and preserves fine-grained spatial details. The MAA loss further incorporates object scale and modality reliability into localization supervision, extending modality awareness beyond feature fusion to localization optimization. Extensive experiments demonstrate the effectiveness of MQCODet on multiple public UAV-based RGB-T benchmarks.
In future work, we will explore the application of MQCODet to a broader range of domains and further improve its deployment efficiency.

\section*{Acknowledgements}

This work was supported by the Frontier Research Fund of the Institute of Optics and Electronics, Chinese Academy of Sciences, under Grant No.~C24K003.

\FloatBarrier
\setlength{\bibsep}{2pt plus 2pt minus 1pt}

\end{document}